%% file: main.tex
\documentclass[11pt, logo, onecolumn, copyright, colorlinks=true, allcolors=blue]{nvidiatechreport}
\usepackage[numbers]{natbib}
\usepackage{nicefrac}
\usepackage{listings}
\usepackage{subcaption}
\usepackage{adjustbox}
\usepackage{xcolor}
\usepackage{soul}
\usepackage{caption}
\usepackage{colortbl}
\usepackage{pdflscape}
\usepackage{xspace}
\usepackage{pgfplots} 
\usepackage[varqu,varl]{inconsolata}   % clean monospace for code
\usepackage{etoolbox}
\usepackage[skins,breakable,listings]{tcolorbox}
\tcbuselibrary{listingsutf8}
\usetikzlibrary{arrows.meta,positioning,calc,fit,backgrounds,decorations.pathreplacing,shapes.geometric}   % tikz itself comes via tcolorbox
\usepackage{hyperref}
\usepackage{fontawesome5}
\usepackage{booktabs}
\usepackage{array}
\usepackage{graphicx}
\usepackage[font=small,labelfont=bf]{caption}

\definecolor{nvgreen}{RGB}{118,185,0}      % NVIDIA-green left accent bar
\definecolor{kwgreen}{RGB}{84,128,18}      % keywords
\definecolor{builtinblue}{RGB}{41,103,160} % builtins / self
\definecolor{strteal}{RGB}{17,133,128}     % strings
\definecolor{commentgray}{RGB}{122,128,118}% comments
\definecolor{codebg}{RGB}{248,249,246}     % panel background
\definecolor{codeframe}{RGB}{224,227,220}  % panel border
\definecolor{ctxslate}{RGB}{92,101,112}    % static/prefix slate accent
\definecolor{ctxamber}{RGB}{206,133,38}    % static-context accent
\definecolor{ctxgray}{RGB}{130,137,145}    % completion / output accent
\definecolor{decoratorfuchsia}{RGB}{205, 50, 160}

\newcommand{\nooa}{\textsc{NOOA}\xspace}
\newcommand{\nooalong}{{NVIDIA Object-Oriented Agents}\xspace}

\newcommand{\legendbox}[2]{{\setlength{\fboxsep}{1.5pt}\colorbox{#1}{#2}}}

\newcommand{\authorsep}{\hspace{0.6em}}
\newcommand{\authorseplong}{\hspace{2em}}

\newcommand{\repolink}{%
  \href{https://github.com/NVIDIA-NeMo/labs-OO-Agents}{%
    {\normalsize
      \textcolor{black}{{\faGithub}}\hspace{0.35em}%
      \textcolor{kwgreen}{\texttt{nvidia-nemo/labs-OO-Agents}}%
    }%
  }%
}

\definecolor{codebackground}{HTML}{F6F8FA} 
\definecolor{codetext}{HTML}{000000} % Pure black text

\newtcbox{\inlinecode}[1][]{
    on line,
    boxrule=0pt,
    colback=codebackground,
    coltext=codetext,
    arc=2pt, % Subtle rounded corner
    before upper={\strut\ttfamily\small}, % Keeps font proportional
    boxsep=0pt,
    left=3pt,  % Snug side padding
    right=3pt,
    top=0pt,   % No extra height at the top
    bottom=0pt,% No extra height at the bottom
    #1
}

\newcommand{\principlesymbol}{%
  {\color{nvgreen}\rule[0.2ex]{0.55em}{0.55em}}%
}

\newcommand{\principle}[2]{%
  \par\smallskip
  \noindent
  \principlesymbol%
  \hspace{0.45em}%
  \textbf{P#1. #2}~%
}

\newtcolorbox{capabilitybox}{
  enhanced,
  colback=nvgreen!12,
  colframe=nvgreen!95,
  boxrule=0.75pt,
  arc=1.5pt,
  left=6pt,
  right=6pt,
  top=4pt,
  bottom=4pt,
  boxsep=0pt,
  before skip=0.4em,
  after skip=0.7em
}

\newcommand{\capabilityitem}[2]{%
  \par\smallskip\noindent
  \hangindent=1.2em
  \hangafter=1
  \textbf{\textsc{#1}.} #2%
}

\newcommand{\capabilities}[1]{%
  \begin{capabilitybox}
  \vspace{-0.15em}
  {\small #1}
  \end{capabilitybox}
}

\newcommand{\inlinecapabilitybox}[1]{%
  \tcbox[
    on line,
    enhanced,
    colback=nvgreen!12,
    colframe=nvgreen!65,
    boxrule=0.45pt,
    arc=1.5pt,
    left=2pt,
    right=2pt,
    top=0.6pt,
    bottom=0.6pt,
    boxsep=0pt
  ]{\small #1}%
}

\newcommand{\inlineparagraph}[1]{%
  \par\medskip
  \noindent\textbf{#1}\hspace{0.5em}%
}

\definecolor{commentgreen}{RGB}{70,120,70}

\newtcolorbox{codebox}{
  enhanced,
  colback=codebg,
  colframe=codeframe,
  boxrule=0.4pt,
  arc=2.5pt,
  leftrule=0pt,
  borderline west={2.2pt}{0pt}{nvgreen},
  boxsep=2pt,
  left=10pt,
  right=8pt,
  top=5pt,
  bottom=5pt,
  before skip=10pt,
  after skip=10pt,
}

\title{\begin{center}
NVIDIA-labs OO Agents\\
Native Python Object-Oriented Agents
\end{center}}

\author{%
\begin{minipage}{\textwidth}
\centering

Paul Furgale\authorseplong
Severin Klingler\authorseplong
James Nolan\authorseplong
Matt Staats
\par\vspace{0.25em}

Gaia Di Lorenzo\authorsep
Elisa Martinez Abad\authorsep
Christian Schüller\authorsep
Razvan Dinu
\par\vspace{0.25em}

Alessio Devoto\authorseplong
Pascal Berard\authorseplong
Gal Kaplun\authorseplong
Elad Sarafian\authorseplong
\par\vspace{0.25em}

Riccardo Roveri\authorseplong
Leon Derczynski\authorseplong
Ricardo Silveira Cabral
\par\vspace{1.4em}

\repolink

\end{minipage}
}
\begin{document}
\begin{abstract}
Traditional agent development is split across prompt templates, tool schemas, callback code, and workflow graphs. We present \nooalong (\nooa or NVIDIA double-O
Agents), a model-agnostic Python framework for building reliable AI agents. \nooa takes a simpler approach: an agent is a Python object. Its methods are
the actions the model can take, fields are its state, docstrings are its prompts, and its type annotations are contracts. A method with code body consisting of \inlinecode{\texttt{...}} is completed at runtime by an LLM-driven agent loop, while methods with normal bodies remain standard deterministic Python. This gives developers and agents the same interface, so agent behavior can be tested, traced, refactored, and improved just like other software.

This paper makes three contributions. 
(1)~We present the \textbf{agent-as-a-Python-object programming model} and the design principles behind it. Where Python has existing abstractions, we adopt them directly: agents are classes, capabilities are methods, type annotations are contracts, asynchronous work is \inlinecode{\texttt{asyncio}}, and tools and orchestration are normal Python code. Agent-specific capabilities -- context, events, state rendering, long-term memory, and validated LLM loops -- are exposed through simple Pythonic APIs, so both developers and agents share one familiar programming model.
(2)~We identify \textbf{six model-facing ideas} that \nooa is, to our knowledge, the first to combine on a single surface: \textit{typed input/output,} \textit{pass-by-reference} over live objects, \textit{code as action}, \textit{programmable loop engineering}, \textit{explicit object state}, and \textit{model-callable harness APIs} for context and events. Surveying fourteen agent frameworks and harnesses, we find the community already converging on several of these ideas -- often as experimental or partial features -- and we present the comparison to encourage further adoption.
(3)~We demonstrate that \textbf{current models use this interface effectively}, both in targeted capability tests and on SWE-bench Verified and Terminal-Bench 2.0; on the ARC-AGI-3 interactive-reasoning benchmark, the interface compresses a multi-agent world-model system into a single agent with a one-page skill while advancing the benchmark's score--cost Pareto frontier.

\end{abstract}
\maketitle

\section{Introduction}
\label{sec:intro}
% ===========================================================================

With the increasing interest in AI agents, there has been a proliferation of agent development kits, each with its own developer-facing and model-facing abstractions~\cite{langchain2023,llamaindex2023,anthropic2024effectiveagents,anthropic2025longrunningharnesses,anthropic2025multiagent}. 
These systems expose useful primitives -- tools, memory, workflows, handoffs, traces, and code execution -- but they often split agent source code across prompt templates, schemas, callbacks, configuration files, and orchestration code. Consequently, learning a new agent framework often means learning a new programming model for capabilities that already have mature equivalents in ordinary programming languages: typed interfaces, variable scoping, control flow, asynchronous execution, and object state. These abstractions are not only familiar to developers, but also broadly represented in model training data.

\nooalong (\nooa) is inspired by PyTorch~\cite{paszke2019pytorch}, which showed that \textbf{a powerful runtime can still present users with a simple Python programming model}. \nooa applies the same concept to agents: where Python already has the right abstraction, \nooa uses it. Agent actions, helper logic, and harness extension points are ordinary Python programs, familiar to developers, close to the distribution of code that LLMs were trained on, and thus directly understandable by coding agents. Where agent-specific concepts do not already have a standard Python form -- for example context construction, event history, and model-visible state -- \nooa exposes them as \textit{simple Pythonic APIs}. This design provides a dual benefit: it eliminates the learning curve for humans and ensures immediate \textbf{agent readiness}.

A complete agent is a single Python class, as shown in the support-agent example below. The class combines object state, deterministic code and two agentic methods: a single-shot \inlinecode{Predict} method and an iterative \inlinecode{CodeAct} method. 
\begin{center}
\begin{minipage}{\linewidth}
\begin{codebox}
\begin{lstlisting}
from nooa import Agent 

TicketKind = Literal["refund", "damaged", "other"]

# Return type: validated by the runtime before triage() returns.
# Descriptions and constraints are model-visible.
class Ticket(BaseModel):
   kind: TicketKind
   priority: int = Field(ge=1, le=5, description="Urgency from 1 (low) to 5 (high).")
   summary: str = Field(description="Customer-visible summary of the issue.")

# The Agent is a Python object.
class SupportAgent(Agent):
   """You are a support agent for a customer service system."""

   # Object state: model-visible, passed by reference.
   order_db: OrderDB

   # Real body: ordinary Python. Deterministic, testable, callable by the model.
   def is_refund_eligible(self, order: Order) -> bool:
       """Return whether an order is eligible for a refund."""
       return order.delivered and order.days_since_delivery <= 30

   # "..." body: an agentic method. Predict makes a single typed LLM call.
   @strategy(PredictStrategy())
   async def classify(self, message: str) -> TicketKind:
       """Classify the customer message into the best ticket kind."""
       ...

   # The default strategy, CodeAct, runs a loop in which the model writes
   # Python: it can inspect order, call is_refund_eligible() and classify(),
   # and must return a Ticket. Inputs are live objects, not serialized text.
   @strategy(CodeActStrategy())
   async def triage(self, message: str, photo: Image | None, order: Order | None) -> Ticket:
       """Triage a customer message and create a support ticket."""
       ...
\end{lstlisting}
\end{codebox}
\captionof{figure}{\textbf{Implementation of a simple Agent in \nooa.}}
\label{lst:support-agent}
\end{minipage}
\end{center}
The class is simultaneously source code, prompt surface, type contract, tool interface, and state boundary. A method with code executes as ordinary Python; a method whose body contains an ellipsis (\inlinecode{\texttt{...}}) becomes an \textit{agentic method}, so the harness runs it as an LLM-driven loop. 

The method declaration specifies the loop: the signature gives the model structured inputs and an output-validation contract, the docstring becomes the prompt, and methods on \inlinecode{\texttt{self}} and imported libraries become callable tools. Note that inputs are not limited to text, as  \inlinecode{\texttt{triage}} receives an image and a live  \inlinecode{\texttt{Order}} object, passed by reference rather than serialized into the prompt. This brings prompt engineering back into software engineering, so behavior can be tested, traced, refactored, versioned, and optimized.

The rest of the paper develops this design. Sec~\ref{sec:design-principles} presents the design principles. Sec~\ref{sec:harness-contributions} shows how they are realized in the programming model and harness. Sec~\ref{sec:evaluation} tests whether current models can use this interface, with capability tests and results on SWE-bench Verified, Terminal-Bench 2.0, and ARC-AGI-3. Sec~\ref{sec:comparison-to-other-agent-development-frameworks} compares fourteen other frameworks and harnesses against the six interface capabilities. Sec~\ref{sec:related} situates the capabilities in the broader literature and Sec~\ref{sec:conclusion} discusses limitations and future work.

% ===========================================================================
\section{Design Principles}
\label{sec:design-principles}
% ===========================================================================
%
Five principles guided the design. Each principle materializes as one or more interface capabilities implemented by \nooa. In the following, we mark each design principle with a square (\principlesymbol), and use a callout to name the corresponding \inlinecapabilitybox{\textbf{interface capabilities}}. The principles state the design commitments behind \nooa, the capabilities name the concrete model-facing features reused throughout the implementation.

\principle{1}{Reuse Python abstractions}
\textit{If a mature Python abstraction already exists, adopt it rather than introducing a domain-specific language (DSL).} In \nooa, classes define agents, methods define capabilities, fields hold explicit, model-visible durable state, type annotations define contracts, \texttt{asyncio} expresses concurrency, exceptions signal failures, and control flow is ordinary Python available to developers and agents alike.
\capabilities{
\capabilityitem{Loop engineering}{Control flow for single and multi-agent orchestration is ordinary Python.}
\capabilityitem{Object state}{Durable state is stored on the agent object, rather than only in conversation history.}
}
\principle{2}{Reframe agentic loops as method calls}
\textit{The application sees an agentic loop as a normal Python method call with typed input/output, not an unstructured text exchange.} Arguments are passed by reference as live Python objects, while the harness renders bounded previews and context to the agent, injects arguments and object state into the loop, and validates return values before returning to the caller.
\capabilities{
\capabilityitem{Typed I/O}{Agentic methods have typed inputs and typed return values.}
\capabilityitem{Pass by reference}{The model operates on live Python objects by reference.}
}
\principle{3}{Move deterministic work out of the agentic loop}
\textit{LLMs are useful for semantic judgment, synthesis, and open-ended tasks. Exact rules, arithmetic, parsing, and state transitions belong in deterministic methods.}
The boundary is local and visible in the code: a real method body for deterministic work, an ellipsis (\inlinecode{\texttt{...}}) body for agentic loops.
\principle{4}{Unlock the model's existing Python knowledge}
\textit{LLMs already know how to write Python and use popular Python libraries.} By letting models write normal Python code instead of tool calls, \nooa draws on that knowledge. CodeAct code can use ordinary loops and conditionals, \inlinecode{\texttt{asyncio}} for concurrency, database clients for queries, plotting libraries for visualization, and ordinary imports for extension -- without bespoke prompting, reading documentation, or learning a new DSL. This makes \nooa exceptionally easy to use while maximizing \textbf{agent readiness}, ensuring that the library is as intuitive for autonomous coding agents to build with as it is for human developers.
\capabilities{
\capabilityitem{Code as action}{The model acts by writing Python code, control flow and method calls directly.}
}
\principle{5}{Expose the harness as explicit APIs}
\textit{Agent-specific concepts -- structured context, context rendering, and event history -- are exposed as Python APIs to developers and the model.} Where possible, the
interfaces mirror built-in types or existing libraries so they are familiar and obvious. The Agent has access to its own context and is able to manage it via Pythonic primitives.
\capabilities{
\capabilityitem{Harness APIs}{Harness and Context are exposed through explicit APIs both to the user and to the agent.}
}
%
%
% These principles produce the six model-facing ideas compared in Sec~\ref{sec:related}: loop engineering and object state (Principle~1), typed input/output and pass by reference (Principle~2), code as action (Principle~4), and model-callable harness APIs (Principle~5).
% ===========================================================================
\section{Agent Loop}
\label{sec:harness-contributions}
A \nooa agent is a Python object that exposes model-callable behavior through typed methods, fields, and docstrings. Developers write and use this object as ordinary Python code. At runtime, the harness executes regular methods directly and implements ellipsis-body methods as LLM loops. This section unrolls that loop: context rendering, pass by reference, Python execution, event and state recording, and return validation.
\subsection{Agents and Strategies}
\label{sec:agent-loop-implementation}
An agent may contain both ordinary Python methods and \textit{agentic} methods -- methods whose body contains the ellipsis literal, \inlinecode{\texttt{...}}. Control flow remains ordinary Python until execution reaches an agentic method; at that point, the harness implements the method as an agent loop. The docstring and method arguments become the prompt for the current task, the type signature defines the input and output contract, and the model may use the methods and state on  \inlinecode{\texttt{self}} before returning the result. The support-agent example in Sec~\ref{sec:intro} shows both kinds of method in one class.

\inlineparagraph{Strategies} \nooa implements agentic methods through \emph{strategies}. A strategy is declared as a decorator: it preserves the method's ordinary Python signature and typed boundary, but controls its agentic execution -- what context is rendered, how turns are executed, and how candidate outputs are validated. Strategies are per-method, and they are an extension point: new strategies can be added as the field progresses. The decorator also takes per-method overrides -- model, truncation, and scoped context -- so, for example, a small fast model can serve a classification method while the agent's default model serves open-ended ones. Within a single agent, externally initiated calls to agentic methods are serialized, so independent invocations do not interleave their turns. Nested same-agent calls follow stack discipline: the caller is suspended until the callee returns, and both executions append to the same event history. Other methods, and other agents, run in parallel under Python's standard async/await concurrency model.
\nooa provides two built-in strategies:
\begin{enumerate}
\item \inlinecode{\texttt{\textbf{PredictStrategy}}} is a single-shot strategy for classification or extraction: it renders the context, asks the model for a value, then validates the output against the Python return type, running a local retry loop if the output fails validation.
\item \inlinecode{\texttt{\textbf{CodeActStrategy}}} generalizes the same contract into an iterative Python Read-Eval-Print Loop (REPL). The model may call \inlinecode{\texttt{execute\_python(...)}} to compute, inspect internal agent state, call helpers, or invoke other generation methods; the harness records the observation, re-renders the updated state, and repeats until the model calls \inlinecode{\texttt{return\_result(...)}} with a value that is type validated.
\end{enumerate}
The same agent can mix both strategies, choosing per method whichever execution mode fits the task: in the support-agent example,  \inlinecode{\texttt{classify\_ticket}} uses Predict and \inlinecode{\texttt{triage}} uses the default CodeAct.
\input{codeact_loop_v2}

Figure~\ref{fig:agent-loop} shows the agent loop for the CodeAct strategy. The rest of this section follows the loop: the harness first \emph{renders context} from the method call (Sec.~\ref{sec:agent-loop-render-context}); it then \emph{calls the LLM} (Sec.~\ref{sec:agent-loop-call-llm}); if the model chooses a code action, the harness \emph{executes Python} in the method's REPL session (Sec.~\ref{sec:agent-loop-execute-python}); finally, it \emph{updates events and state} with the code output, errors, return values, and locals before the next turn is rendered (Sec.~\ref{sec:agent-loop-update-events-state}). When the model submits a result, the harness validates it against the return type (Sec.~\ref{sec:agent-loop-strong-typing}); failures return an error message to the model, and success returns control to the caller.
\subsection{Context}
\label{sec:agent-loop-render-context}
\begin{figure}[tbp]
\centering
\begin{tikzpicture}[
  >={Triangle[]}, font=\footnotesize,
  mgr/.style 2 args={draw=#1, line width=0.9pt, rounded corners=4pt,
        top color=white, bottom color=#1!12, align=center,
        text width=#2, minimum height=1.4cm, inner sep=4pt},
  region/.style 2 args={draw=#1, line width=0.9pt, rounded corners=3pt,
        top color=white, bottom color=#1!14, align=center, text width=#2,
        minimum height=2cm, inner sep=4pt, font=\bfseries, text=black},
  up/.style={-{Triangle[length=2.6mm,width=2.3mm]}, line width=1pt, #1},
  up/.default=black!55,
  badge/.style={draw=black!35, fill=black!5, rounded corners=2pt,
        inner xsep=3pt, inner ysep=1.2pt, minimum height=3.4mm,
        text height=1.35ex, text depth=.35ex, font=\tiny\ttfamily, text=black!60},
]
  % ---- the three regions of the context window ----------------------------
  \node[region={ctxamber}{0.27\linewidth}]   (rs) at (-0.29\linewidth,0)
        {Static Context\\[3pt]{\scriptsize\normalfont\linespread{0.85}\selectfont static instructions visible at every turn.\par}};
  \node[region={builtinblue}{0.27\linewidth}] (re) at (0,0)
        {Events\\[3pt]{\scriptsize\normalfont\linespread{0.85}\selectfont the agent's execution history.\par}};
  \node[region={kwgreen}{0.27\linewidth}]     (rd) at (0.29\linewidth,0)
        {Dynamic Context\\[3pt]{\scriptsize\normalfont\linespread{0.85}\selectfont per-turn rendered state or environment information.\par}};

  % ---- role badges: pinned to the bottom-right edge of each region --------
  \node[badge, anchor=north east] at ([yshift=-1.5pt]rs.south east) {system};
  \node[badge, anchor=north east] (et) at ([yshift=-1.5pt]re.south east) {tool\_call};
  \node[badge, anchor=north east] (ea) at ([xshift=-3pt]et.north west)   {assistant};
  \node[badge, anchor=north east]        at ([xshift=-3pt]ea.north west) {user};
  \node[badge, anchor=north east] at ([yshift=-1.5pt]rd.south east) {user};

  % ---- managers: the programmable sources ---------------------------------
  \node[mgr={ctxamber}{0.34\linewidth}]   (cm) at (-0.22\linewidth,3.0)
        {\textbf{ContextManager}\\[2pt]{\scriptsize context blocks: static + dynamic}};
  \node[mgr={builtinblue}{0.30\linewidth}] (em) at ( 0.25\linewidth,3.0)
        {\textbf{EventManager}\\[2pt]{\scriptsize typed events (FIFO queue)}};

  % rounded right-angle routing (single clean crossing in the middle)
  \draw[up=ctxamber,    rounded corners=6pt] (cm.south) -- ++(0,-0.55) -| (rs.north);
  \draw[up=ctxamber,    rounded corners=6pt] (cm.south) -- ++(0,-0.55) -| (rd.north);
  \draw[up=builtinblue, rounded corners=6pt] (em.south) -- ++(0,-0.30) -| (re.north);
\end{tikzpicture}

% ---- the rendered XML for each region (row 4) -----------------------------
\vspace{1mm}
\begin{center}
\begin{minipage}[t]{0.27\linewidth}
\begin{tcblisting}{enhanced, colback=codebg, colframe=codeframe,
  boxrule=0.35pt, arc=2pt, left=2pt, right=2pt, top=2pt, bottom=2pt,
  equal height group=rendercontextsnippets, valign=top, before skip=0pt, after skip=0pt,
  listing only, listing options={language=XML,basicstyle=\ttfamily\tiny,
  breaklines=true, columns=fullflexible, showstringspaces=false}}
<system_prompt>
  You are TaskDecomposition
  Agent, a Python agent
  working in an interactive
  session...
</system_prompt>
<self expr="doc(type(self))">
  class TaskDecompositionAgent:
    async def get_users(
      self) -> list[dict]:
      """Get the users from
      the database."""
</self>
\end{tcblisting}
\end{minipage}\hspace{0.02\linewidth}%
\begin{minipage}[t]{0.27\linewidth}
\begin{tcblisting}{enhanced, colback=codebg, colframe=codeframe,
  boxrule=0.35pt, arc=2pt, left=2pt, right=2pt, top=2pt, bottom=2pt,
  equal height group=rendercontextsnippets, valign=top, before skip=0pt, after skip=0pt,
  listing only, listing options={language=XML,basicstyle=\ttfamily\tiny,
  breaklines=true, columns=fullflexible, showstringspaces=false}}
<sys tag="1">
  Task(prompt="## Task:
   parse_normalize_and_
   validate_records ...")
</sys>
<tool_call
 name="execute_python">
  reasoning("Inspecting
    inputs...")
  pprint(tasks, max_depth=4)
</tool_call>
<sys tag="3">
  PythonOutput(stdout="tasks
   (list): ['Convert the
   *created* date ...']")
</sys>
\end{tcblisting}
\end{minipage}\hspace{0.02\linewidth}%
\begin{minipage}[t]{0.27\linewidth}
\begin{tcblisting}{enhanced, colback=codebg, colframe=codeframe,
  boxrule=0.35pt, arc=2pt, left=2pt, right=2pt, top=2pt, bottom=2pt,
  equal height group=rendercontextsnippets, valign=top, before skip=0pt, after skip=0pt,
  listing only, listing options={language=XML,basicstyle=\ttfamily\tiny,
  breaklines=true, columns=fullflexible, showstringspaces=false}}
<context>
  <todos
   expr="self.todo.status()">
    1/3 done
    [x] parse created date
    [ ] parse bought date
    [ ] clean descriptions
  </todos>
</context>
\end{tcblisting}
\end{minipage}
\end{center}
\caption{\textbf{Context rendering in \nooa.} The  \inlinecode{\texttt{ContextManager}} and  \inlinecode{\texttt{EventManager}} populate static
context, event history, and dynamic context before each LLM turn.}
\label{fig:render-context}
\end{figure}
% \subsubsection{Context is a Programmable Object}
% \label{sec:render-context-programmable-object}
%
%
The first step in a CodeAct turn is to render the live Python execution state into model context. \nooa separates context into three regions (see Figure~\ref{fig:render-context}): \textit{static context blocks}, which are computed once and reused across turns; \textit{event history}, which records the execution trace accumulated so far; and \textit{dynamic context blocks}, which are re-evaluated before each model call.

\inlineparagraph{Static and Dynamic Blocks} These are developer-controlled, named, structured pieces of text rendered into the model's context window. Static blocks hold information that stays stable across the call, such as the system prompt. Dynamic blocks hold information whose value changes as the program runs, such as a \verb|TODO list| or selected relevant fields on  \inlinecode{\texttt{self}}.

\inlineparagraph{Event History} This is an append-only sequence of typed events produced by the harness as execution proceeds: model tool calls, Python outputs, and return values. Each event is a typed Python object with a unique tag, so agent code can query prior events rather than scanning a flat transcript. Long histories can be collapsed into summary events, akin to MemGPT's context management~\cite{packer2023memgpt}; strategies can restrict which events are visible to a nested call, and the full event history remains searchable after summarization. Together, blocks and events form the model context of an agentic method.

Context management is therefore not an external prompt-building script; it is part of the same object-oriented API used by the agent. Both the developer and the agent can interact with the context through Pythonic APIs, as shown in Fig.~\ref{fig:context-apis}.
\begin{center}
\begin{minipage}{\linewidth}
\begin{codebox}
\begin{lstlisting}
# Static context blocks are simple key/value pairs.
self.context["notes"] = "The user wants concise responses."

# Dynamic context blocks accept Python expressions evaluated every turn.
self.context.set_dynamic("todo", "self.todo.status()")

# Event history: query or compact the execution trace.
recent_python = self.events.query(type="PythonOutput", limit=3)
self.events.collapse(start_tag, end_tag, summary_text="Model generated summary.")
\end{lstlisting}
\end{codebox}
\captionof{figure}{\textbf{Context engineering in \nooa.} Context blocks and the event history are Python APIs available to both the developer and the agent.}
\label{fig:context-apis}
\end{minipage}
\end{center}
\nooa starts with defaults that make simple agents work well, while still allowing developers to dynamically override every context block at any time. The default static prefix contains a small \nooa system prompt (about 1k characters), the active strategy instructions (about 2.5k characters for CodeAct), an execution-context block showing imported types and libraries, and a concise  \inlinecode{\texttt{doc(self)}} rendering of the agent API. The dynamic suffix contains compact views of live agent state (\inlinecode{\texttt{pprint(self)}}). The helper \inlinecode{\texttt{doc()}} provides documentation for types, while \inlinecode{\texttt{pprint()}} formats values and instances. Unless scoped by a strategy or method, the event-history block renders the visible execution events accumulated so far.

\inlineparagraph{Rendering context} These three sources are maintained by two programmable objects, shown in Figure~\ref{fig:render-context}: the  \inlinecode{\texttt{ContextManager}}, which stores static and dynamic \emph{context blocks}, and the \inlinecode{\texttt{EventManager}}, which stores the \emph{event history} as an ordered log of typed events.
The renderer maps these sources into LLM API messages (e.g., OpenAI chat messages). Static framework blocks, such as \inlinecode{\texttt{<system\_prompt>}} and \inlinecode{\texttt{<self>}} (the agent's own  \inlinecode{\texttt{doc()}} rendering), are concatenated into a cacheable \emph{system} prefix visible at every turn. The event history becomes the interleaved \emph{user}, \emph{assistant}, and \emph{tool} messages that record execution: system-generated task messages, agent \inlinecode{\texttt{tool\_call}} s, and Python output. Dynamic blocks are re-rendered every turn into a trailing \emph{user}  \inlinecode{\texttt{<context>}} message. Each dynamic block shows its expression to the model (e.g., \inlinecode{\texttt{expr="self.todo.status()"}}), reinforcing that this is live state. This three-region layout is designed to maximize KV-cache reuse across turns: the static prefix remains unchanged, the event history grows only by appending new messages, and volatile dynamic blocks are placed at the tail. As a result, updates to live state do not invalidate the cached prefix, and each turn can reuse most of the previous computation.

By default, context blocks and events are wrapped in XML-like tags and events are rendered as typed Python \inlinecode{\texttt{repr}}s, as shown at the bottom of Figure~\ref{fig:render-context}. Media arguments -- images, audio, video, and files -- are rendered as native multimodal content blocks rather than text, which is how \inlinecode{\texttt{triage}} in the intro example receives its \inlinecode{\texttt{photo}}. The renderer is an extension point: developers have full control over what goes in the context and how it is rendered.

\inlineparagraph{Pass by Reference} Rendering context does not mean serializing the whole program state into the prompt. A CodeAct method receives its arguments as live Python objects, and for large arguments the model never sees the full value. In the spirit of progressive disclosure~\cite{anthropic2025agentskills}, the model sees each argument's \emph{variable name} paired with a bounded preview: the concrete type, the true length, and a short head/tail sample. The model reads that shape, understands that the name refers to a real object, and operates on it directly in generated code.

For example, a method called with a list of one hundred integers renders in the prompt as a single compact preview:
\begin{codebox}
\begin{lstlisting}[language=Python]
records = list(len=100, [:5]=[42, 17, 89, 33, 8], [-5:]=[56, 71, 12, 45, 28])
\end{lstlisting}
\end{codebox}

\noindent The preview states the concrete type (\inlinecode{\texttt{list}}), the true length (\inlinecode{\texttt{len=100}}), and a head/tail sample; the elided middle is implied.  The variable \inlinecode{\texttt{records}} itself is \emph{not} truncated -- it is the full hundred-element list bound as a local in the execution environment -- so the model can index, slice, or iterate over all of it (\inlinecode{\texttt{for r in records: ...}}) even though only ten elements ever appear in the context window.

This is what lets the object model scale past the context window: the amount of data an agent can process is bounded by the execution environment, not by the prompt.  A method can accept a multi-million-row table or a multi-megabyte string and the agent works on the whole thing by writing code, while the prompt carries only a fixed-size preview.

Python has no standard library for truncating arbitrary values. The closest is Rich's \inlinecode{\texttt{pprint()}}~\cite{rich}, so we borrowed its name and API surface -- both are in the model's training data -- but changed the output format based on experimentation across open and closed models. Finding even better formats that are obvious to LLMs, and supporting more types, remains open work.

Methods using the Predict strategy render argument values in full, guarded by a size cap: a Predict call is a single LLM call, so the model has no opportunity to inspect a variable.

\subsection{Calling the LLM}
\label{sec:agent-loop-call-llm}
Once the harness has rendered the current turn, control passes from Python to the model. The LLM receives the structured context assembled in the previous step, together with the strategy-specific contract for what it may do next. Under \inlinecode{\texttt{PredictStrategy}}, the model must produce a value matching the return annotation. Under \inlinecode{\texttt{CodeActStrategy}}, the model must choose between continuing computation with \inlinecode{\texttt{execute\_python(...)}} or terminating the method with \inlinecode{\texttt{return\_result(...)}}.

\subsection{Executing Python}
\label{sec:agent-loop-execute-python}

When a CodeAct model chooses a Python action, \nooa executes the cell in a restricted, Jupyter-like session. Method arguments, the live agent as \inlinecode{\texttt{self}}, and the agent's environment (imports, methods, and constants defined in the agent's source file) are injected as locals; \inlinecode{\texttt{await}} can be used directly. The cell can inspect objects with \inlinecode{\texttt{doc(obj)}}, print bounded previews with \inlinecode{\texttt{pprint()}}, call deterministic helpers, await generation methods, spawn subagents, or return an in-process Python value with \inlinecode{\texttt{return\_result(...)}}.

This is the second half of pass by reference mentioned in~Sec~\ref{sec:agent-loop-render-context}: the model writes code against real objects rather than serialized tool arguments. All tool calls are strongly typed and pass by reference in both directions, so the agent can call a method with a huge input, bind the huge typed result to a variable, and process it programmatically -- slice it, aggregate it, feed it to the next call -- while only the bounded previews it chooses to print enter the context window. Models already improvise this pattern in bash -- spilling results to files and processing them with follow-up commands; \nooa replaces the untyped text on disk with typed, live variables that persist from cell to cell.

Dangerous or loop-breaking APIs such as \inlinecode{\texttt{eval}}, \inlinecode{\texttt{exec}}, \inlinecode{\texttt{compile}}, \inlinecode{\texttt{input}}, and blocking event-loop calls are rejected with specific errors. Stdout, stderr, images, returned values, locals, and exceptions are captured as structured results. Syntax errors and tracebacks are in IPython format, including source locations and caret/source-line context, so the next LLM turn can repair the code the way a human would repair a notebook cell.

Cells can contain loops, conditionals, library calls, async operations, helper calls, and subagent invocations. This gives the model the same orchestration tools as the developer: inside a cell, it can define a new \inlinecode{\texttt{@strategy}}-decorated function with an ellipsis body and fan it out over a batch with \inlinecode{\texttt{asyncio.gather}}, creating parallel subagent calls in ordinary Python. 

\subsection{Updating Events and State}
\label{sec:agent-loop-update-events-state}
After every model response or Python execution, the harness appends typed events to the event manager: tool calls, Python outputs, and final return values.

State updates follow standard Python scoping rules. REPL locals are method-scoped -- they persist across cells within a single CodeAct call and then disappear when the method returns -- so intermediate values stay local to the task. Anything reached through \inlinecode{\texttt{self}} or through library calls, by contrast, can have side effects that outlive the method, exactly as they would in an ordinary Python program.

\subsection{Validating the Return}
\label{sec:agent-loop-strong-typing}
When the model returns a result, the harness validates it against the return annotation. If the result is invalid, the harness sends the model an error message describing the failure, and the loop continues. If the result is valid, the harness returns it to the caller and normal Python execution resumes.

\input{memory_system}
% ===========================================================================

\section{Evaluation}
\label{sec:evaluation}
We evaluate \nooa at two levels. First, in Sec.~\ref{sec:evaluation:do-models-understand-the-harness}, we use targeted capability tests to determine whether current models understand and correctly use the abstractions exposed by the \nooa interface. Second, we evaluate complete \nooa agents end-to-end on benchmarks spanning software engineering and terminal interaction (Sec.~\ref{sec:evaluation:is-the-interface-effective}), cybersecurity (Sec.~\ref{sec:cybergym}), and interactive reasoning (Sec.~\ref{sec:arc3}).
\subsection{Capability Tests: Do Models Understand the \nooa Interface?}
\label{sec:evaluation:do-models-understand-the-harness}

\inlineparagraph{Experimental setup}
We built a suite of focused integration tests that isolate one interface behavior at a time.  The question is not only whether a model can solve a task, but whether it can call helper methods, write executable cells, interpret bounded variable previews, manage state, and return typed values through the harness. The suite contains 88 test instances across 36 families, covering typed method calls, structured returns, stateful object manipulation, routing to helper agents, context and truncation handling, REPL and code execution, batching through generated loops, error recovery, and task decomposition.

Most tests are short interactions of one to five turns. The harder cases stress bookkeeping over batches, recovery after errors, multi-step REPL exploration, and the implementation of reusable helper methods. The complete suite is included in the \nooa repository. We run each test five times for each of ten models, yielding 4,400 records in total.

\inlineparagraph{Models understand the interface} Table~\ref{tab:captests-model-summary} shows that current generation models are generally
fluent in the \nooa interface; the suite passes 4,309 of 4,400
records (97.9\%). We group models by scale: four small/efficient models (Claude Haiku 4.5,
Gemini 3.5 Flash, Nemotron 3 Nano 30B, GPT-5.4 Mini) and six large/frontier models
(Claude Opus 4.8, Gemini 3.1 Pro, GLM-5.2, Kimi K2.6, Nemotron 3 Ultra, GPT-5.5).
Small/efficient models pass 96.0\% of records; large/frontier models pass 99.2\%.
Every model exceeds 91\%, and six of ten models exceed 98\%. 
GPT-5.5 is perfect on this suite, while Gemini 3.5 Flash and GLM-5.2 miss only one test each. Capability-suite pass rates discriminated by the use of reasoning show frontier models saturating regardless of mode (Opus 100.0/99.5, GPT-5.5 99.5/98.6, off/on), while the value of reasoning grows monotonically as model capability falls — Ultra 93.4 → 94.1, Super-v3 83.7 → 96.4, Nano 52.5 → 84.8 — making inference-time reasoning a capability equalizer for the smaller Nemotron models.
\begin{table}[t]
\centering
\caption{Capability-test pass rates. Each model is evaluated on 440 records: 88 tests, five runs each.}
\label{tab:captests-model-summary}
\small
\begin{tabular}{lrr}
\toprule
\textbf{Model} & \textbf{Passed} & \textbf{Pass rate} \\
\midrule
Claude Haiku 4.5 & 430/440 & 97.7\% \\
Claude Opus 4.8 & 438/440 & 99.5\% \\
Gemini 3.5 Flash & 439/440 & 99.8\% \\
Gemini 3.1 Pro (preview) & 438/440 & 99.5\% \\
GLM-5.2 & 439/440 & 99.8\% \\
Kimi K2.6 & 431/440 & 98.0\% \\
Nemotron 3 Nano 30B & 403/440 & 91.6\% \\
Nemotron 3 Ultra & 434/440 & 98.6\% \\
GPT-5.4 Mini & 417/440 & 94.8\% \\
GPT-5.5 & 440/440 & 100.0\% \\
\midrule
\textbf{Overall} & \textbf{4309/4400} & \textbf{97.9\%} \\
\bottomrule
\end{tabular}
\end{table}
The important implication is that the interface itself is not a burden for current generation LLMs. Models know Python; they can read object documentation, call methods with typed arguments, use returned values, mutate object state, and return values that satisfy the type contract. This zero-shot fluency validates the framework's empirical agent readiness: by expressing agentic constructs as native software abstractions, we completely remove the interface friction introduced by other frameworks. 
\inlineparagraph{Stress tests expose the remaining frontier}
 The residual failures are concentrated in six stress families, shown in
 Table~\ref{tab:captests-stress}. These are the tests that most resemble
 agentic work rather than single tool calls: preserving per-item bookkeeping in a
 large batch, recovering from errors, iterating in a REPL, refining an intermediate
 answer, and decomposing repeated transformations into helpers. The stress subset
 passes 254 of 300 records (84.7\%), compared with 97.9\% overall. Large/frontier
 models pass 169 of 180 stress records (93.9\%), while small/efficient models pass
 85 of 120 (70.8\%) -- the scale gap widens from 3.2 points overall to 23 points
 on the stress subset.

Running each test five times also measures consistency. Models are consistent: of the 880 (test, model) pairs, 94\% pass all five runs, only three fail all five, and the rest are intermittent. The stress tests separate the two failure modes: large models have no 0/5 scores -- every failure is intermittent, a reliability miss on a demonstrated capability. Small models show both, with 12.5\% of stress pairs at 0/5 and 42\% intermittent. 

\begin{table}[t]
\centering
\caption{Stress-test pass rates. Each row has 50 records
(10 models $\times$ five runs), split into four small/efficient and six large/frontier models as defined in the text.}
\label{tab:captests-stress}
\scalebox{0.88}{%
\begin{tabular}{lrrr}
\toprule
\textbf{Stress test} & \textbf{Small/efficient} & \textbf{Large/frontier} & \textbf{Overall} \\
\midrule
\texttt{sentiment\_batch} & 8/20 (40\%) & 23/30 (76.7\%) & 31/50 (62\%) \\
\texttt{calculate\_batch} & 14/20 (70\%) & 27/30 (90\%) & 41/50 (82\%) \\
\texttt{refinement} & 11/20 (55\%) & 30/30 (100\%) & 41/50 (82\%) \\
\texttt{task\_decomposition} & 15/20 (75\%) & 30/30 (100\%) & 45/50 (90\%) \\
\texttt{error\_recovery} & 19/20 (95\%) & 29/30 (96.7\%) & 48/50 (96\%) \\
\texttt{repl\_exploration} & 18/20 (90\%) & 30/30 (100\%) & 48/50 (96\%) \\
\midrule
\textbf{Stress aggregate} & \textbf{85/120 (70.8\%)} & \textbf{169/180 (93.9\%)} & \textbf{254/300 (84.7\%)} \\
\bottomrule
\end{tabular}
}
\end{table}

These are not failures to understand \inlinecode{\texttt{self}} or to call a method; they are failures of disciplined multi-step harness use (Appendix~\ref{app:sentiment-batch-traces} shows four complete runs of the hardest stress test). This distinction matters: basic interface fluency is already widespread, while reliable long-horizon batching, recovery, and decomposition at the code/model interface remain capability frontiers. 
\subsection*{Experimental Results on Agentic Benchmarks}
\label{sec:exp_results_benchmarks}
We evaluate \nooa on four agent benchmarks covering complementary forms of end-to-end interaction. SWE-bench Verified~\cite{jimenez2024swebench} measures software-engineering performance on real repository issues, while Terminal-Bench 2.0~\cite{tbench_2025} evaluates multi-step interaction in a command-line environment. CyberGym L1~\cite{wang2026cybergym} tests an agent's ability to identify and repair software vulnerabilities, and ARC-AGI-3~\cite{arcprize2026arcagi3} evaluates interactive reasoning in unfamiliar environments. Together, these benchmarks span code modification, terminal use, cybersecurity, and adaptive problem solving.
\subsection{Software Engineering and Terminal Interaction}
\label{sec:evaluation:is-the-interface-effective}
\input{swe_results_v2}
\subsection{Securing Software on CyberGym L1}
\label{sec:cybergym}
\input{cybergym}
\subsection{Advancing the score--cost Pareto frontier on ARC-AGI-3}
\label{sec:arc3}
\input{arc_agi_3_results}
\section{Comparison to other harness libraries}
\label{sec:comparison-to-other-agent-development-frameworks}
In Sec~\ref{sec:design-principles}, we identified six interface capabilities that \nooa combines: \textit{typed I/O, pass by reference, code as action, programmable loop engineering, object state}, and \textit{model-visible harness APIs}. The previous sections showed how these capabilities are implemented in \nooa and how they affect agent behavior. We now compare \nooa with fourteen agent frameworks and harnesses along the same axes. The comparison shows that prior systems support important subsets of these capabilities, but, to the best of our knowledge, \nooa is the first agent development kit to expose all six on a single surface. Table~\ref{tab:harness-idea-comparison} provides an overview.

\inlineparagraph{How we scored the results.}
We scored each system by reading its documentation and source code. Every score is checked against a pinned snapshot; the repository, commit, and package version are listed in each system's subsection of Appendix~\ref{app:harness-comparison-evidence} (snapshots retrieved July 7--9, 2026). Green (\emph{Supported}) means the capability is a first-class part of what the model sees. Yellow (\emph{Partial}) means it exists, but mainly for the developer, or behind a tool or a file. Red (\emph{Limited}) means we found no evidence of it. Experimental, flag-gated, or opt-in capabilities are scored on the capability itself and marked \dag{} rather than demoted. For harness APIs the bar is that the model itself can see or call the context and event machinery; tracing dashboards, automatic compaction, and hidden callbacks do not count. Table~\ref{tab:harness-idea-comparison} gives the scores with a short reason per cell; Appendix~\ref{app:harness-comparison-evidence} has the full evidence. 

\textbf{No other system combines all six ideas, but most are adopting some of them}. Most systems have a version of each idea, but expose it to the developer instead of the model, or wrap it in a new abstraction where a mature one already exists. The newest, strongest capabilities -- Microsoft's harness providers, Pydantic's CodeMode harness, OpenAI's sandbox agents, Codex's code mode -- shipped during our evaluation window, most marked experimental or flag-gated (\dag). We read this as the field converging on these six ideas.
\begin{table*}[!tp]
 \centering
 \caption{
 Emerging design patterns across agent development kits and harnesses.
 The field is broadly converging on six harness-interface patterns; shading
 shows how each system realizes each pattern today:
 \legendbox{green!22}{native}, \legendbox{green!8}{emerging}, or
 \legendbox{black!5}{minimal}. Detailed evidence appears in
 Appendix~\ref{app:harness-comparison-evidence}.
 }
 \label{tab:harness-idea-comparison}
 \newcolumntype{L}{>{\raggedright\arraybackslash}p{0.17\linewidth}}
 \newcolumntype{C}{>{\centering\arraybackslash}p{0.125\linewidth}}
 \newcommand{\nativebox}{\cellcolor{green!22}}
 \newcommand{\partialbox}{\cellcolor{green!8}}
 \newcommand{\earlybox}{\cellcolor{black!5}}
 \newcommand{\strongcell}[1]{\nativebox\strut #1}
 \newcommand{\partialcell}[1]{\partialbox\strut #1}
 \newcommand{\limitedcell}[1]{\earlybox\strut #1}
 \setlength{\tabcolsep}{3.5pt}
 \renewcommand{\arraystretch}{1.24}
 {\sffamily\scriptsize
 \scalebox{0.92}{%
 \begin{tabular}{LCCCCCC}
 \toprule
 System
 & \shortstack{Typed I/O}
 & \shortstack{Pass by\\ reference}
 & \shortstack{Code as action}
 & \shortstack{Loop\\ engineering}
 & \shortstack{Object state}
 & \shortstack{Harness APIs} \\
 \midrule
   LangGraph / LangChain
   & \partialcell{typed output only}
   & \partialcell{tool-side state/ store}
   & \partialcell{shell tool\textsuperscript{\dag}}
   & \partialcell{dev: graph DSL}
   & \partialcell{todo/ memory tools\textsuperscript{\dag}}
   & \partialcell{model: tool-search loading\textsuperscript{\dag}} \\
   LangChain Deep Agents
   & \partialcell{typed output only}
   & \partialcell{files in graph state}
   & \strongcell{shell tool\textsuperscript{\dag}; JS REPL\textsuperscript{\dag}}
   & \strongcell{dev: graph DSL; model: delegation tool, JS loops\textsuperscript{\dag}}
   & \partialcell{todo list; memory files\textsuperscript{\dag}}
   & \partialcell{model: compaction tool\textsuperscript{\dag}} \\
   Microsoft Agent Framework
   & \partialcell{typed output only}
   & \partialcell{file mounts \& tools\textsuperscript{\dag}}
   & \strongcell{code cell, inline tools\textsuperscript{\dag}}
   & \partialcell{dev: workflow DSL; model: background agents\textsuperscript{\dag}}
   & \partialcell{todo/ memory/ mode tools\textsuperscript{\dag}}
   & \partialcell{model: memory/ todo tools\textsuperscript{\dag}} \\
   OpenAI Agents SDK\textsuperscript{\ddag}
   & \partialcell{typed output only}
   & \partialcell{container files\textsuperscript{\dag}}
   & \partialcell{code \& shell tools}
   & \partialcell{dev: SDK; model: handoffs}
   & \limitedcell{host-side sessions}
   & \partialcell{model: skill/ tool-search loading} \\
   Google ADK
   & \strongcell{typed input \& output (agent-as-tool)}
   & \partialcell{named artifacts}
   & \partialcell{code executor (no tools inside)}
   & \partialcell{dev: workflow DSL; model: transfer/ exit tools}
   & \partialcell{templated state; memory tool}
   & \partialcell{model: artifact/ memory loading} \\
   PydanticAI
   & \partialcell{typed output only}
   & \partialcell{sandbox variables\textsuperscript{\dag}}
   & \strongcell{code cell, inline tools\textsuperscript{\dag}}
   & \partialcell{dev: typed graphs; model: delegation tools}
   & \partialcell{provider memory tool\textsuperscript{\dag}}
   & \partialcell{model: capability/ tool-search loading} \\
   smolagents
   & \partialcell{objects in/ out, untyped}
   & \strongcell{live objects in namespace}
   & \strongcell{code executor}
   & \strongcell{dev: Python; model: code loops over subagents}
   & \partialcell{untyped state dict; persistent namespace}
   & \limitedcell{dev hooks only} \\
   Claude Agent SDK
   & \partialcell{typed output only}
   & \partialcell{workspace files}
   & \partialcell{shell tool}
   & \strongcell{dev: SDK, hooks; model: subagents, workflow scripts}
   & \partialcell{model-edited memory files; task list}
   & \partialcell{model: skill/ tool-search loading} \\
   OpenAI Codex
   & \partialcell{typed output only}
   & \partialcell{workspace files \& MCP resources}
   & \strongcell{shell tool; code cell\textsuperscript{\dag}}
   & \strongcell{dev: SDK, CI; model: delegation tools, code loops\textsuperscript{\dag}}
   & \partialcell{plan/ goal tools; memory files\textsuperscript{\dag}}
   & \partialcell{model: skill/ tool-search loading} \\
   OpenHands
   & \limitedcell{text in/ text out}
   & \partialcell{workspace files}
   & \partialcell{shell tool; workflow cell\textsuperscript{\dag}}
   & \strongcell{dev: SDK; model: workflow scripts\textsuperscript{\dag}; delegation tool\textsuperscript{\dag}}
   & \partialcell{todo tool; plan files}
   & \partialcell{model: skill loading} \\
   PI
   & \limitedcell{text in/ text out}
   & \partialcell{files}
   & \partialcell{shell tool}
   & \partialcell{dev: TS SDK; model: delegation tool\textsuperscript{\dag}}
   & \limitedcell{host-side sessions}
   & \partialcell{model: skill loading} \\
   Hermes
   & \limitedcell{text in/ text out}
   & \partialcell{files w/ preview \& paging}
   & \strongcell{code cell, inline tools}
   & \partialcell{model: parallel delegation}
   & \partialcell{model-edited memory files}
   & \partialcell{model: memory/ session search} \\
   OpenCode
   & \limitedcell{text in/ text out}
   & \partialcell{files}
   & \strongcell{shell tool; code cell (MCP tools)\textsuperscript{\dag}}
   & \partialcell{dev: JS SDK; model: delegation tool}
   & \limitedcell{write-only todos}
   & \partialcell{model: skill loading} \\
   OpenClaw
   & \limitedcell{text in/ text out}
   & \partialcell{shared workspace files}
   & \strongcell{shell tool, or code-mode cell\textsuperscript{\dag}}
   & \strongcell{model: session spawn/ send; code spawn loops\textsuperscript{\dag}}
   & \partialcell{model-edited memory files}
   & \partialcell{model: session/ memory tools} \\
 \bottomrule \addlinespace[2pt]
 \multicolumn{7}{@{}l@{}}{\scriptsize\textsuperscript{\dag}\,available as an extension, flag-gated, or opt-in --- not enabled by default.} \\
 \multicolumn{7}{@{}l@{}}{\scriptsize\textsuperscript{\ddag}\,scored on the core SDK; the beta sandbox subsystem adds model-visible memory files.} \\
 \end{tabular}
 }
 }
\end{table*}
\section{Related Work}
\label{sec:related}
The comparison in Sec~\ref{sec:comparison-to-other-agent-development-frameworks} evaluates existing harnesses against the six interface capabilities implemented by \nooa. This section situates those capabilities in the broader literature. We group prior work by the contribution to agent harness design: structured and typed LLM programming, executable code as an action interface, programmable orchestration, state and memory, and model-visible harness operations.

% ===========================================================================
\inlineparagraph{Typed I/O:} Agent calls have typed inputs and a typed return value. 
DSPy~\cite{khattab2024dspy} made declarative signatures -- named input and output fields, with data types added in later releases -- the unit of LLM programming, decoupling declared intent from prompt wording and making pipelines programmatically optimizable. LMQL~\cite{beurer2023lmql} frames prompting as a query language whose output constraints (including type constraints) are enforced at decoding time, and Outlines~\cite{willard2023outlines} enforces output structure during generation by compiling regular expressions to finite-state machines (and grammars to pushdown automata) that mask invalid tokens at each step. Engineering libraries such as Instructor~\cite{instructor2023} and TypeChat~\cite{typechat2023} validate model \emph{output} against a schema and re-prompt with the validation errors.  Mainstream agent frameworks have converged on the same need: LangChain agents~\cite{langchain2023} and PydanticAI~\cite{pydanticai2024} accept an output schema, and Google's ADK~\cite{adk2025} enforces an output schema on agent replies and additionally accepts an input schema when an agent is exposed as a tool via its agent-as-tool path. 

Agentic methods and tools in \nooa are defined as Python methods, and type annotations at generation-method boundaries are enforced by the runtime.

\textbf{Code as action:} The model acts by writing arbitrary code.
PAL~\cite{gao2022pal} and Program of Thoughts~\cite{chen2022pot} offload computation to generated programs, and Chain of Code~\cite{li2023chaincode} interleaves interpreter execution with LM-simulated execution of the lines an interpreter cannot run. CodeAct~\cite{wang2024codeact} consolidates the argument that executable code should be the action modality itself, outperforming JSON and text actions; smolagents~\cite{smolagents2024} packages the paradigm as a library, executing model-written code in a restricted interpreter with tools as callable functions; and OpenHands~\cite{openhands2024}, built on CodeAct, demonstrated the paradigm at scale on software engineering tasks (its V1 SDK rebuild has since moved to discrete shell and file tools; see Sec~\ref{sec:comparison-to-other-agent-development-frameworks}) (SWE-Agent~\cite{yang2024sweagent} showed, complementarily, that the agent--computer interface itself is a first-class design surface). Anthropic's programmatic tool calling has the model invoke tools as functions from within an executing program rather than as one JSON call per turn, keeping bulk intermediate results out of the context window~\cite{anthropic2025advancedtooluse}. Recursive Language Models~\cite{zhang2025recursive} and Recursive Agent Harnesses~\cite{lumer2026recursive} study long-context decomposition through recursive model or harness calls.

Bash tools are themselves a weak form of code as action: a shell command line is a small program, with pipes and loops for control flow and CLIs as callable tools. We believe this explains the rise of tool-as-CLI over tool-as-MCP packaging -- a CLI is called from code, so the model can filter, transform, and compose outputs programmatically, while an MCP tool is a single JSON call whose full result lands in the context window. The shell's limitations remain: it operates only on untyped text, with no variable persistence outside of files.

Recent survey work frames the same shift more broadly as \emph{code as agent harness}~\cite{ning2026codeharness}: code becomes the substrate for reasoning, acting, environment modeling, execution-based verification, planning, memory, tool use, and multi-agent coordination. \nooa has realized this design as an object-oriented Python runtime.

\textbf{Pass by reference:} The model operates on live, in-process objects. 
Most agent frameworks use copy-as-text at every interface. Inputs are serialized to text, tool call inputs are generated as text by the LLM and outputs are returned as text, and finally LLM output text is parsed back into the host language~\cite{openai_function_calling}. Cheng et~al.~\cite{cheng2025sharing} propose \emph{shared program state} as a natural function interface, allowing prompts to read and write live program state via explicit references. Their programming system Nightjar embeds natural-language code blocks inside Python programs, using angle-bracket notation (\inlinecode{\texttt{<var>}} to reference and \inlinecode{\texttt{<:var>}} to assign shared variables) and an interface through which the LLM manipulates state.
AskIt~\cite{okuda2024askit} provides a type-guided domain specific language that turns typed prompt templates into callable functions, serializing captured host variables into the prompt and parsing typed output back. ANPL~\cite{anpl2023} interleaves user-written Python-like sketches with LLM-implemented natural-language holes. CodeAct~\cite{wang2024codeact} replaces the structured tool-calling format with a Python REPL in which tools are ordinary functions and live objects persist across turns; \nooa's default strategy descends from this paradigm, as does smolagents~\cite{smolagents2024}, which injects developer-supplied Python objects into the executor namespace for the model to use by name; Recursive Language Models~\cite{zhang2025recursive} push reference-passing to its logical conclusion: the prompt itself becomes a variable in a REPL that the model inspects, slices, and recursively queries with sub-model calls rather than reading it in full.  TaskWeaver~\cite{taskweaver2024} maintains cross-step state as live Python variables. 

Many popular frameworks today use files as a variant of pass by reference, giving the agent the filename as input and allowing it to explore it via tool calls. This is powerful, but it loses all type information and requires the agent to reconstruct any types embedded in the file by reading the text. 
 
When \nooa operates in CodeAct mode, inputs and tool arguments are live Python objects in the session namespace, and computed outputs are returned by reference from inside code (\inlinecode{\texttt{return\_result(variable)}}). Every call begins with an input-inspection step that prints each parameter's type and a size-bounded preview; the live variable remains in the session for further exploration. This keeps context usage small and under the agent's control, and it allows processing of large inputs by shape without the full content ever entering the context window.

\textbf{Loop engineering:} Control flow for single- and multi-agent orchestration is available to developers and the agent.
Many popular agent-building frameworks provide developer APIs for orchestrating agents: LangGraph~\cite{langgraph2024} expresses control flow as a graph of nodes that read and write typed shared state, and Microsoft Agent Framework~\cite{maf2025} builds workflows as directed graphs of executors exchanging typed messages. Google's ADK~\cite{adk2025} pairs developer-defined workflow agents with model-callable control primitives. smolagents injects developer-defined managed agents into the model's code namespace, so the model can write loops over subagents in its own actions. Claude Code's dynamic workflows have the model itself author the orchestration code: given a task, the agent writes a script that fans out tens to hundreds of parallel subagents with verification stages before results are returned~\cite{anthropic2026dynamicworkflows}. \nooa needs no separate workflow language: outer loops are ordinary Python methods, inner loops belong to the model, and the agent can write and invoke the same control flow the developer does.

\textbf{Object state:} The agent has explicit, model-visible durable state.
For most agents, the conversation history \emph{is} the state. As conversations get long and history is compacted, state can be lost. 
MemGPT~\cite{packer2023memgpt} treats the LLM as an operating system that pages information between in-context and external memory tiers. They reserve a fixed-size read/write ``working context'' section of the prompt that is exempt from eviction, so key facts survive long conversations. 

Today's harnesses keep durable state in three ways: model-edited files (memory files like \texttt{MEMORY.md} re-injected at session start, todo and plan stores, \texttt{AGENTS.md} instructions), searchable conversation history (full-text or vector search over past session transcripts), and dedicated memory tools with add/replace/remove verbs over a store (Sec~\ref{sec:comparison-to-other-agent-development-frameworks}). All three survive compaction and sessions, but all three hold untyped text outside the model's working state.

\nooa's state is object-scoped and part of the contract: typed fields and named context blocks live on the agent instance, and public fields are rendered into the prompt from the live object each turn -- held out of history eviction rather than reconstructed from a transcript.

\textbf{Harness APIs:} Context blocks, per-turn dynamic context, and event inspection are exposed as \emph{model-callable} APIs.
MemGPT and its successor Letta~\cite{packer2023memgpt,letta2024} expose the harness itself as model-callable tools: the model edits its own in-context memory blocks, searches its message history, and pages file content in and out of context -- though context compaction remains harness-triggered, with the model only warned of memory pressure. Memory-R1~\cite{yan2025memoryr1} trains such capabilities with reinforcement learning: a memory-manager agent learns add/update/delete/no-op operations over a memory store while a separate answer agent learns to filter and reason over retrieved memories. \nooa exposes its harness uniformly through the object model: static and dynamic context blocks and the queryable event history are model-callable APIs whose visibility to the model the developer opts into per agent, with scoped overrides available to harness code. We believe that giving agents direct, programmatic access to their context and history will be key to unlocking new agentic behavior on long-running tasks.

\section{Conclusion}
 \label{sec:conclusion}
 % ===========================================================================

\nooa brings together many of the recent advances in agent design into a single ergonomic
development kit in which an agent is a Python object with methods and state.
This design makes agentic software ordinary software: developers and agents use the same interface, the same libraries, and the same tools. Our evaluation shows that current models already operate this interface effectively, despite never being trained on it.

\textbf{Limitations:} \nooa executes model-written code in the agent's own process. The validator in Sec~\ref{sec:agent-loop-execute-python} protects the agent loop, not the host. In this respect \nooa's isolation philosophy is the same as any harness with a shell tool: sandboxing -- a container, VM, or permission system -- goes around the agent process, and a shell tool is no safer than in-process Python; most harnesses in Sec~\ref{sec:related} ship one. Executing in-process is what preserves pass by reference; sandboxed code modes trade it away, receiving serialized copies at the sandbox boundary. Our preferred deployment is OpenShell~\cite{nvidia2026openshell}.

There are several promising directions to pursue:

\textbf{Agent optimization via agent rewriting:}
First, agent optimization should move
beyond prompt search toward rewriting every part of the agent: prompts,
docstrings, typed method signatures, helper code, tool descriptions, context
policies, retry loops, and decomposition structure.
GEPA-style reflective optimization is a natural starting point~\cite{agrawal2026gepa},
but the richer target is the whole agent object and its
harness~\cite{agrawal2026gepa}.

\textbf{Skills as full software packages:}
Second, typed interfaces and libraries create a
path toward self-evolving agents. Today's skills are often text snippets or
informal procedures; we expect skills to become full software libraries with typed
APIs, documentation, tests, examples, subagents, dependencies, and versioned
interfaces that agents can inspect, call, repair, and extend.

\textbf{Reinforcement learning to unlock inductive reasoning:}
Third, reinforcement learning can target the inductive reasoning needed by
long-running agents. DeepSeek-R1 showed that outcome-driven reinforcement
learning can induce useful reasoning behaviors when a model is allowed to search
through intermediate reasoning steps~\cite{deepseekai2025r1}.
We expect a similar effect for object-oriented agents, but over a richer action space than
text alone. A \nooa agent can choose what context to reveal, which variables to
preserve, when to write deterministic helper code, when to promote a pattern
into a reusable library, and when to decompose a task into deterministic orchestration. These are
inductive decisions: the agent must generalize from prior trajectories to new
tasks by identifying which abstractions, state variables, and decomposition
strategies predict success. The hypothesis is that reinforcement learning over
complete agent trajectories could teach models to use harness APIs, dynamic
context, pass-by-reference objects, and code-as-interface as a learned reasoning
substrate. In this view, the harness is not merely an execution environment; it
is the action space in which agents learn to construct, test, and reuse
problem-solving structure.

Taken together, these directions suggest that progress in agent capability will come not only from larger models or better prompts, but from the co-development of model and harness. We believe the software interface is the right place for that co-development. \nooa is one step toward it: an object-oriented harness in which agents are programs that both humans and models can read, execute, test, and improve.

% ===========================================================================
% References
% ===========================================================================

\bibliographystyle{plainnat}
\bibliography{references}

\appendix
\section{Appendix: Harness comparison details}
\label{app:harness-comparison-evidence}

This appendix expands the compact comparison in Table~\ref{tab:harness-idea-comparison}. Each subsection begins with the pinned source snapshot (repository, commit, and package version, all retrieved on July~7, 2026) against which the scores were verified. Scores use the paper's model-visible rubrics: typed loop I/O requires typed inputs and outputs at the model-facing loop; pass-by-reference requires live object references rather than serialized text; code as action requires executable code with control flow and inline tool or method calls; loop engineering asks whether developers and models can program orchestration loops; object state asks whether the model can store and retrieve state through its working interface -- Supported requires typed, model-visible state that is live within the session, so append-only memory text applied at the next session, and memory reachable only through dedicated tools, score Partial; and Harness APIs asks whether structured context blocks, per-turn dynamic context, and session events are exposed as model-visible APIs rather than hidden host machinery.

\subsection{LangGraph / LangChain}
\begin{description}
  \item[Evidence base.] \texttt{langchain-ai/langgraph} at commit \\ \texttt{23652c54be18ce59f697aa38f10075ee91913220} (langgraph 1.2.8) and \\ \texttt{langchain-ai/langchain} at commit \\\texttt{2d8100c4faef2e4a0ec7ab74536fe7a9d9ae551e} (langchain 1.3.11, including the \texttt{langchain\_v1} agent stack).
  \item[Typed loop I/O.] Partial. \texttt{StateGraph} supports typed graph state, input, and output, and \texttt{create\_agent} returns a typed \texttt{structured\_response} when given a \texttt{response\_format}; but the model-facing loop remains messages-in/tool-calls-out -- the typed boundary is the developer-authored graph, not a typed model-facing method call.
  \item[Pass-by-reference.] Partial. Live state and store objects can be injected into model-invoked tool calls (\texttt{InjectedState}, \texttt{InjectedStore}, \texttt{ToolRuntime}), but these arguments are deliberately excluded from the schemas shown to the model -- the model never holds or names a live reference; file-oriented middleware surfaces path/text handles only.
  \item[Code as action.] Partial. The default action modality is JSON tool calling; shell and code execution exist only as opt-in middleware (\texttt{ShellToolMiddleware} with host/sandbox/Docker execution policies) or provider-native pass-through. Neither repository contains a CodeAct-style loop.
  \item[Loop engineering.] Partial. Developers get conditional edges, \texttt{Send} fan-out, and \texttt{Command} control flow; nothing model-side ships in these repositories -- the same-vendor Deep Agents harness (scored separately below) adds tool-mediated subagent dispatch.
  \item[Object state.] Partial. Graph state, stores, reducers, and checkpointers are developer/node state; model-visible durable state exists only through opt-in tools -- e.g., \texttt{TodoListMiddleware}'s \texttt{write\_todos}, the Anthropic memory and text-editor middleware backed by virtual files in graph state, and \texttt{FilesystemFileSearchMiddleware}'s search tools (with a state-backed variant in the Anthropic partner package).
  \item[Harness APIs.] Partial. The v1 middleware stack gives developers rich per-turn request mutation (\texttt{dynamic\_prompt}, \texttt{wrap\_model\_call}, context editing, summarization) -- developer hooks and automatic compaction, which do not qualify. One opt-in surface is model-callable\textsuperscript{\dag}: \texttt{ProviderToolSearchMiddleware} defers selected tool schemas and gives the model a provider-native search tool that loads them on demand. Loading only; no model-callable context blocks or event inspection.
\end{description}

\subsection{LangChain Deep Agents}
\begin{description}
  \item[Evidence base.] \texttt{langchain-ai/deepagents} at commit \\\texttt{7be76c752117e6e61dcdc931ea5147261fad6768} (deepagents 0.6.12).
  \item[Typed loop I/O.] Partial. \texttt{create\_deep\_agent} takes a first-class \texttt{response\_format} passed through to \texttt{create\_agent}, and the compiled graph's output state carries a typed \\ \texttt{structured\_response}; subagents may declare their own \texttt{response\_format}, serialized as JSON into the \texttt{task} tool result. Input remains untyped chat messages (plus a \texttt{files} dict), so typing covers output only.
  \item[Pass-by-reference.] Partial. A virtual filesystem lives in graph state (\texttt{FilesystemState.files}) behind pluggable backends (state, disk, store, composite, sandbox). Oversized tool results are evicted to \texttt{/large\_tool\_results/...} paths the model re-reads with \texttt{read\_file}, and subagent file writes merge back into parent state. All indirection is by file-path string; no live typed object handles.
  \item[Code as action.] Strong (opt-in). The first-party \texttt{langchain-quickjs} partner package adds a persistent JS REPL \texttt{eval} tool -- a code cell with control flow and inline \texttt{task} subagent dispatch (programmatic tool calls are a further opt-in), vendor-marked beta\textsuperscript{\dag}. In the core package the action modality is JSON tool calling: an \texttt{execute} shell tool is registered in the built-in suite but functions only when the backend implements the sandbox protocol; with the default state backend it is withheld from the model request\textsuperscript{\dag}.
  \item[Loop engineering.] Strong (opt-in). Developers get the full langgraph graph and a composable middleware stack. By default the model gets delegation only: the \texttt{task} tool launches ephemeral, stateless subagents (parallel fan-out encouraged), and opt-in async subagents add start/check/update/cancel/list tools for graphs served via the Agent Protocol\textsuperscript{\dag}. The opt-in beta JS REPL lets the model author orchestration loops in code, with inline \texttt{task} calls\textsuperscript{\dag}.
  \item[Object state.] Partial. Model-writable durable state is planning- and file-shaped: \texttt{write\_todos} maintains a typed todo list in session state, virtual files persist in state, \texttt{memory=} loads \texttt{AGENTS.md} files into the system prompt with the model persisting learnings via \texttt{edit\_file}, and a store backend adds cross-thread durability. No typed model-visible live objects beyond the todo schema.
  \item[Harness APIs.] Partial. By default summarization is automatic, skills are prompt-injected metadata the model expands via \texttt{read\_file}, and memory (when configured) loads automatically. An opt-in middleware exposes a model-callable \texttt{compact\_conversation} tool\textsuperscript{\dag} -- one manipulation verb, short of the model-callable context-block, dynamic-context, and event surface this axis requires for Supported -- and evicted history is re-openable at a known path via \texttt{read\_file}.
\end{description}

\subsection{Microsoft Agent Framework}
\begin{description}
  \item[Evidence base.] \texttt{microsoft/agent-framework} at commit \\ \texttt{9c4cd07899502157284b64a73f9a0adfb4594d96} (agent-framework-core 1.10.0; Python packages). Several harness-like capabilities on this commit are gated \texttt{@experimental} and are noted as such.
  \item[Typed loop I/O.] Partial. Workflows are typed (\texttt{WorkflowContext}) and agent responses are generically typed via \texttt{response\_format} (\texttt{AgentResponse[T].value}), but run input is untyped messages and \texttt{agent.as\_tool()} reduces a subagent to a single string \texttt{task} argument -- the model-facing loop contract stays text-mediated.
  \item[Pass-by-reference.] Partial. Files are the reference substrate: CodeAct file mounts with model-visible instructions and returned file artifacts, plus model-callable \\ \texttt{file\_access\_write}/\texttt{file\_access\_read} tools (\texttt{FileAccessProvider}). Live objects never cross the boundary -- the CodeAct bridge rejects any non-JSON-safe value at the sandbox/host boundary.
  \item[Code as action.] Strong (opt-in). The Monty/Hyperlight CodeAct providers make model-written code the sole action surface when enabled: provider-owned tools are hidden from the model and invoked from inside code -- through generated, type-checked stubs (Monty) or an untyped \texttt{call\_tool} built-in (Hyperlight). It is an opt-in provider package, not the default modality; both packages are prerelease (Monty alpha, Hyperlight beta), though neither carries the framework's \texttt{@experimental} gate.
  \item[Loop engineering.] Partial. Developers get \texttt{WorkflowBuilder} and orchestration packages; the model gets experimental \texttt{BackgroundAgentsProvider} tools to start, await, list, and iteratively continue concurrent subagent tasks, and CodeAct plus agents-as-tools lets it fan out with \texttt{asyncio.gather}. The background-agent tools are delegation (the model fills parameters), and agents-as-CodeAct-tools is a developer-wired composition not demonstrated in the repository, keeping this short of Supported.
  \item[Object state.] Partial. Experimental harness providers expose model-callable durable state: \texttt{TodoProvider} (add/complete/query todos), \texttt{FileMemoryProvider} (memory-file read/write with an auto-injected index), and \texttt{AgentModeProvider} (\texttt{mode\_set}/\texttt{mode\_get}). This is tool-mediated session/file state, not live object fields.
  \item[Harness APIs.] Partial. \texttt{ContextProvider.before\_run}/\texttt{after\_run} composes per-turn instructions, messages, and tools (developer-authored), and the experimental memory/todo/mode tools are model-callable and change what is injected on subsequent turns; event inspection remains developer-only (workflow events, devui, observability).
\end{description}

\subsection{OpenAI Agents SDK}
\begin{description}
  \item[Evidence base.] \texttt{openai/openai-agents-python} at commit \\ \texttt{078a28f11e8d8f618e08b6c1cd5acf7647137612} (openai-agents 0.18.0). Scores cover the core SDK; this version also ships a beta sandbox-agents subsystem, which is noted where it would change a score.
  \item[Typed loop I/O.] Partial. \texttt{output\_type} gives a typed final output and tools have typed schemas, but run input remains item/message mediated, and even \texttt{Agent.as\_tool(parameters=...)} renders its typed input back into a text preamble for the callee.
  \item[Pass-by-reference.] Partial. The SDK is explicit that local context objects ``are not passed to the LLM''; nested agent-tool results exist only host-side, and the model receives a stringified result. The file substrate arrives with the opt-in container-backed \texttt{ShellTool}\textsuperscript{\dag}: developer files mount into the container (\texttt{file\_ids}), the model addresses them by path, and \texttt{container\_reference} reuses a container across runs. File handles, never live objects.
  \item[Code as action.] Partial. Hosted \texttt{CodeInterpreterTool} and container-mode \texttt{ShellTool}, the executor-backed \texttt{LocalShellTool}, and \texttt{ApplyPatchTool} provide opt-in code execution; ordinary agent actions remain item/tool mediated. In the beta sandbox-agent harness, arbitrary shell becomes the primary action modality, delivered as function-tool calls.
  \item[Loop engineering.] Partial. The model's orchestration verbs are handoffs (\texttt{transfer\_to\_*}) and agents-as-tools; it can iterate by re-calling agent tools but has no API to author orchestration control flow. Developer-side composition happens around \texttt{Runner}.
  \item[Object state.] Limited (core SDK). Sessions are host-side transcript stores with no model-callable tools, and \texttt{RunContextWrapper} is explicitly not model-visible. The beta sandbox Memory capability (model-searchable \texttt{MEMORY.md} with live-update mode) would merit Partial if beta features were scored. (Container persistence via \texttt{container\_reference} is a raw filesystem, credited under pass-by-reference, not a state feature.)
  \item[Harness APIs.] Partial. Two core, GA-documented surfaces are model-callable loading: \texttt{ShellTool} skill mounting (local, inline, and referenced skills rendered into the model-facing payload) and \texttt{ToolSearchTool}, which lets the model search tools marked \texttt{defer\_loading}. Instructions-as-callables and hooks/tracing remain developer-side. The beta sandbox adds a model-callable \texttt{load\_skill} progressive-disclosure tool and injected memory summaries.
\end{description}

\subsection{Google ADK}
\begin{description}
  \item[Evidence base.] \texttt{google/adk-python} at commit \texttt{44d747ed5eaf543b5b8d22e0088f8a7c7eeee846} (google-adk 2.3.0).
  \item[Typed loop I/O.] Strong. \texttt{output\_schema} is enforced at the model boundary (as a response schema, via \texttt{SetModelResponseTool}, or as a typed \texttt{finish\_task}); \texttt{input\_schema} is pydantic-validated where an agent is invoked as a tool or workflow node (\texttt{AgentTool}, \texttt{NodeTool}). The agent-as-tool qualifier is load-bearing: a root agent's user-facing input is untyped.
  \item[Pass-by-reference.] Partial. Artifacts are durable named handles: the model calls \\ \texttt{load\_artifacts} by name, contents are temporarily inserted and removed, and code execution saves its output files back as named artifacts (input files come from inline request data cached in session state). Everything crossing the model boundary is serialized; there are no live objects.
  \item[Code as action.] Partial. A \texttt{code\_executor} on an \texttt{LlmAgent} executes model-emitted code blocks with input files and stdout/stderr/output-artifact observations, across several executors (local, container, GKE, Vertex-family sandboxes, and Gemini's server-side \texttt{BuiltInCodeExecutor}). Executed code is pure compute -- it cannot call tools or agents -- which is the half of this axis's definition (inline tool or method calls) that ADK does not meet.
  \item[Loop engineering.] Partial. Developers get a workflow-graph package (\texttt{Workflow}), with the older \texttt{LoopAgent}/\texttt{SequentialAgent}/\texttt{ParallelAgent} shells deprecated in its favor at this pin; the model gets control primitives (\texttt{transfer\_to\_agent}, \texttt{exit\_loop}) and can invoke developer-authored workflow nodes as tools (\texttt{NodeTool}), but cannot author orchestration loops -- and sandboxed code has no agent access.
  \item[Object state.] Partial. Session state is templated into instructions by default \\ (\texttt{\{var\}}, \texttt{\{artifact.file\_name\}}), \texttt{output\_key} writes model replies into durable state,\\ \texttt{load\_memory} is model-callable memory search (\texttt{preload\_memory} injects memory automatically, without a model call), and tools mutate \texttt{ctx.state} directly. The model cannot enumerate or write state fields directly.
  \item[Harness APIs.] Partial. \texttt{load\_artifacts}, \texttt{load\_memory}, \texttt{load\_mcp\_resource}, and a four-tool skills surface (list and load skills and their resources, run skill scripts; a fifth search tool appears with a registry) are model-callable dynamic-context loading APIs. Context assembly is otherwise developer-mediated: \texttt{ToolContext} is systematically hidden from model-facing schemas, compaction (when configured; itself experimental) runs automatically without model involvement, and there is no model-callable event inspection.
\end{description}

\subsection{PydanticAI}
\begin{description}
  \item[Evidence base.] \texttt{pydantic/pydantic-ai} at commit \texttt{fc597c480abe04b81857bddd0e832adc4c8c896a} (retrieved 2026-07-08; the package version is dynamic from git). Main package \\ \texttt{pydantic\_ai\_slim/pydantic\_ai/}, plus \texttt{pydantic\_graph/}.
  \item[Typed loop I/O.] Partial. The typed-output story is the strongest among the frameworks compared: \texttt{Agent} is generic in its output type, \texttt{output\_type} accepts Pydantic models, unions, and typed output functions, enforced via four modes (\texttt{ToolOutput}, \texttt{NativeOutput}, \texttt{PromptedOutput}, \texttt{TextOutput}), and validation failures or \texttt{ModelRetry} are re-prompted to the model. Input, however, remains untyped: \texttt{run(user\_prompt: str | Sequence[UserContent] | None)}; \texttt{deps\_type} is developer-side injection, and there is no built-in agent-as-tool exposing a typed input schema -- delegation is a developer-written tool function. Typed output only.
  \item[Pass-by-reference.] Partial\textsuperscript{\dag}. Live objects (\texttt{deps}) are injected only into developer-written functions (tools, instructions, validators) via \texttt{RunContext}; the model can never name or hold a reference to them, and multimodal inputs (\texttt{BinaryContent}, \texttt{FileUrl}) are developer-passed content. The credit comes from the official \texttt{pydantic-ai-harness} add-on's CodeMode\textsuperscript{\dag}: sandbox variables persist across \texttt{run\_code} calls, so the model can bind a tool result to a name and pass it to the next tool without the value transiting its context. Values still cross the sandbox boundary as serialized plain data -- tool results are dumped to JSON-compatible form before entering the Monty sandbox, and \texttt{run\_code} accepts no variable injection (verified at \texttt{pydantic/pydantic-ai-harness} commit \texttt{b4365440}). Copies, never live host objects.
  \item[Code as action.] Strong (opt-in). The default action modality is JSON tool calls, but the harness add-on's CodeMode makes model-written code the action surface when enabled\textsuperscript{\dag}: registered tools collapse into a single \texttt{run\_code} tool and are called inline from the model's code as typed Python functions, statically type-checked against generated stubs, in a persistent Monty REPL. This is the same shape as Microsoft's Monty CodeAct provider and is scored the same. The harness package is itself prerelease (alpha classifier), but CodeMode sits outside its \texttt{experimental} subtree and is vendor-designated as released. The provider-native \texttt{CodeExecutionTool} (Anthropic, OpenAI Responses, Google, Bedrock Nova, xAI) and the optional \texttt{mcp-run-python} sandboxed-Python MCP server also exist\textsuperscript{\dag}.
  \item[Loop engineering.] Partial. Developer-side control is first-class: the typed \texttt{pydantic\_graph} graph library (\texttt{Graph[StateT, DepsT, InputT, OutputT]}), node-level iteration via \\ \texttt{agent.iter()}, plain-Python composition, and durable execution backends. The model does not author orchestration loops: delegation and hand-off are developer-written tool functions, application code, or the harness add-on's experimental string-in/string-out \texttt{delegate\_task} tool\textsuperscript{\dag}; the CodeMode add-on\textsuperscript{\dag} lets the model write multi-tool control flow but is a separate gated package.
  \item[Object state.] Partial. Dependencies are typed but developer-side, and \texttt{message\_history} is developer-managed. Model-visible durable state is the Anthropic-only \texttt{MemoryTool} native tool\textsuperscript{\dag}, which upgrades a developer-supplied \texttt{memory} backend to the provider's native memory tool, plus the harness add-on's experimental \texttt{Planning} capability, whose typed \texttt{write\_plan} tool re-surfaces the current plan each request\textsuperscript{\dag} -- memory- and plan-tool abstractions rather than typed in-session state.
  \item[Harness APIs.] Partial. Core ships two model-callable dynamic-context loading surfaces: the reserved \texttt{load\_capability} tool, through which the model pulls deferred capability bundles (instructions, tools, settings) from a catalog appended to its instructions, and tool search (\texttt{search\_tools} or provider-native), which lets the model discover tools marked \texttt{defer\_loading=True}. As with Codex and the Claude Agent SDK, this is context \emph{loading}: there are no model-callable context blocks and no event inspection, so it does not reach Supported.
\end{description}

\subsection{smolagents}
\begin{description}
  \item[Evidence base.] \texttt{huggingface/smolagents} at commit \\ \texttt{526069c1ead958b36d9fd09a6b1ef37f68ed6ade} (smolagents 1.27.0.dev0).
  \item[Typed loop I/O.] Partial. Arbitrarily rich Python values can flow in (\texttt{additional\_args}) and out (\texttt{final\_answer}), but the boundary declares no interface: \texttt{run(task: str)} takes a plain string, the final answer is declared as type \texttt{"any"}, and nothing validates what the model returns; \texttt{final\_answer\_checks} are developer-supplied validator callables, not a model-facing type contract. Individual tools carry typed input schemas and an optional output schema, but these type the tool boundary rather than the loop. The opt-in \texttt{use\_structured\_outputs\_internally} flag\textsuperscript{\dag} constrains only the intermediate thought/code JSON, not the answer type.
  \item[Pass-by-reference.] Strong. \texttt{run(..., additional\_args=\{...\})} places live Python objects (images, dataframes) into \texttt{agent.state}, which is injected verbatim into the code executor's namespace (\texttt{send\_variables}); the task is annotated to tell the model it can access them by key as variables, and the model operates on the actual objects in its generated code. Live objects flow onward to managed agents, and even the JSON-based \texttt{ToolCallingAgent} substitutes state keys in tool arguments with the referenced object. Remote executors (e2b, Docker, Modal, Blaxel) fall back to by-value serialization; reference semantics hold for the default local executor. There is no shaped preview mechanism: the model sees the arguments' untruncated \texttt{str()} appended to the task, plus print output (50k-character cap) and a truncated last-expression value.
  \item[Code as action.] Strong. In \texttt{CodeAgent}, the flagship paradigm, code is the action modality itself: the model's output is parsed as a code blob and executed by a persistent Python executor, with tools and managed agents injected as callable Python functions; interpreter state persists between steps, and observations return execution logs plus the last expression value. Execution targets a restricted local AST interpreter by default or remote sandboxes. \texttt{ToolCallingAgent} provides the JSON alternative.
  \item[Loop engineering.] Strong. Developer-side, orchestration is ordinary Python around \texttt{agent.run()}, plus step callbacks and harness-scheduled re-planning. Model-side, managed agents are injected into the executor namespace as callables alongside tools, and the system prompt instructs the model to call team members like tools -- so the model can author loops, conditionals, and data flow over sub-agents inside its own code action. Unlike \nooa, the model composes developer-defined agents rather than defining new ones; managed agents are unsupported with remote executors.
  \item[Object state.] Partial. \texttt{agent.state} is an untyped \texttt{dict[str, Any]} injected into the executor, and the executor namespace persists across steps and across \texttt{run(reset=False)} calls, so the model holds live objects in session -- stronger than a file substrate. But the state carries no types or schema and is never rendered into context: the model sees only \texttt{str()} of the arguments appended to the task plus whatever it prints, and \texttt{agent.memory.steps} is edited via developer callbacks; the model sees it only as rendered messages, not as typed state.
  \item[Harness APIs.] Limited. Nothing is model-callable for context, dynamic context, or events: planning steps are harness-triggered on a fixed interval, and step callbacks, final-answer checks, \texttt{interrupt()}, and memory editing/replay are all developer-side. No default tool lets the model read its own history or manage its context.
\end{description}

\subsection{Claude Agent SDK}
\begin{description}
  \item[Evidence base.] \texttt{anthropics/claude-agent-sdk-python} at commit \\ \texttt{638e190a91779778a4cd2b00223ce3fd5ad83ae2} (claude-agent-sdk 0.2.111, pinning the Claude Code CLI 2.1.202, fetched at wheel build). The CLI engine is not open source; runtime behavior is verified against Claude Code documentation as of July 2026.
  \item[Typed loop I/O.] Partial. \texttt{query()} and \texttt{ClaudeSDKClient} accept untyped text prompts, but \texttt{output\_format} accepts a JSON schema and the schema-conforming result returns as \texttt{ResultMessage.structured\_output}; custom \texttt{@tool} input schemas are the industry baseline.
  \item[Pass-by-reference.] Partial. The model's reference substrate is the workspace filesystem (paths in prompts, \texttt{CLAUDE.md} \texttt{@path} imports) plus URI-addressed MCP resources; no live object handles cross the model boundary.
  \item[Code as action.] Partial. Actions execute through the \texttt{Bash} tool; there is no code cell with inline tool calls, and the model-written workflow script is restricted to orchestration (no direct filesystem or shell access).
  \item[Loop engineering.] Strong. Developers define subagents, hooks on the SDK's ten lifecycle events, and interrupt and opt-in file-checkpoint rewind controls; the model spawns subagents via the \texttt{Agent} tool and, via the \texttt{Workflow} tool (dynamic workflows, enabled by default on the SDK surface), writes a JavaScript orchestration script whose \texttt{agent()}/\texttt{pipeline()}/\texttt{parallel()} calls fan out up to 16 concurrent and 1{,}000 total subagents with intermediate results held in script variables.
  \item[Object state.] Partial. Auto memory (\texttt{MEMORY.md} plus topic files) is model-written and re-injected each session, \texttt{CLAUDE.md} is developer-authored, subagents can hold persistent memory, and \texttt{TaskCreate}/\texttt{TaskUpdate} maintain a session task list -- file- and tool-mediated rather than typed live state.
  \item[Harness APIs.] Partial. The \texttt{Skill} tool loads skill instructions on demand and \texttt{ToolSearch} loads deferred tool schemas -- both model-callable; compaction is automatic, \texttt{PreCompact} is a developer hook, and context-usage introspection is developer-side only.
\end{description}

\subsection{OpenAI Codex}
\begin{description}
  \item[Evidence base.] \texttt{openai/codex} at commit \texttt{f659eb12bc8cecb976d92db192d9b2983c8053ff} (engine \texttt{codex-rs} plus TypeScript/Python SDKs; in-repo versions are development placeholders, with npm \texttt{@openai/codex} at 0.142.x--0.143.x in July 2026).
  \item[Typed loop I/O.] Partial. \texttt{Thread.run()} accepts prose text (plus local images) and returns typed thread items and a final response; a per-turn \texttt{outputSchema} (JSON Schema) constrains the final message, exposed in both SDKs and as the stable \texttt{--output-schema} flag on \texttt{codex exec}. Structured output is first-class, but input remains free text and the schema-conforming result is returned as an untyped string.
  \item[Pass-by-reference.] Partial. The reference substrate is files and URIs: workspace files, \texttt{AGENTS.md} docs concatenated from project root to cwd, skills advertised by source locator, MCP resources by URI, and images by path via \texttt{view\_image}. There is no non-file mechanism for passing structured data into the loop.
  \item[Code as action.] Strong (flag-gated). ``Code mode'' runs model-authored JavaScript in a V8 runtime with enabled tools bound on a global \texttt{tools} object -- a genuine code cell with control flow and inline tool calls -- but it is under development and off by default\textsuperscript{\dag}. The default action surface is a unified PTY-backed exec tool pair (\texttt{exec\_command}/\texttt{write\_stdin}, stable and default-on except on Windows) plus the \texttt{apply\_patch} structured-diff tool.
  \item[Loop engineering.] Strong (flag-gated). Developers script loops via SDK threads \\ (\texttt{startThread}/\texttt{resumeThread}), \texttt{codex exec} automation, and the external \texttt{openai/codex-action} CI action. The model gets a stable, default-enabled delegation family (\texttt{spawn\_agent}, \texttt{send\_input}, \texttt{resume\_agent}, \texttt{wait\_agent}, \texttt{close\_agent}; a v2 variant replaces these with \texttt{spawn\_agent}, \texttt{send\_message}, \texttt{followup\_task}, \texttt{wait\_agent}, \texttt{interrupt\_agent}, \texttt{list\_agents}), configurable per thread. In the default configuration the model spawns and steers subagents but does not author loops; with code mode enabled the delegation tools bind into the code cell (they pass its exposure filter, and no namespace is excluded by default), so model-written JavaScript can loop over \texttt{spawn\_agent}/\texttt{wait\_agent}\textsuperscript{\dag}. Multi-agent v2 and CSV fan-out jobs are separately flag-gated\textsuperscript{\dag}.
  \item[Object state.] Partial. Durable state is tool- and file-mediated: an always-on \texttt{update\_plan} tool with typed step/status items, persisted per-thread goals via \texttt{get\_goal}/\texttt{create\_goal}/\texttt{update\_goal} (stable, default-on), \texttt{AGENTS.md} files, and an experimental file-backed memories pipeline\textsuperscript{\dag} that extracts memories in the background and injects them at session start. No typed object state lives directly in the model's working interface.
  \item[Harness APIs.] Partial. Model-callable context loading is first-class and on by default for supporting models and providers: \texttt{tool\_search} (BM25 over deferred tool metadata), \texttt{skills.list}/\texttt{skills.read}, and \texttt{list\_mcp\_resources}/\texttt{read\_mcp\_resource}. But there are no model-callable context blocks or event inspection; context-window tools \\(\texttt{get\_context\_remaining}, \texttt{new\_context}) are flag-gated\textsuperscript{\dag}, and lifecycle hooks and auto-compaction remain developer-side.
\end{description}

\subsection{OpenHands}
\begin{description}
  \item[Evidence base.] \texttt{All-Hands-AI/OpenHands} at commit \\ \texttt{1869baf49914309f2115cd8f75d5c7a57a92b371} (openhands-ai 1.10.0) plus \\\texttt{All-Hands-AI/agent-sdk} at commit \\ \texttt{2eff609f9cd4ae31b219d36bd429f74f5348aeef} (openhands-sdk; the app release pins 1.33.0, main is 1.34.0), both retrieved 2026-07-09. The app repo is app/server layers only; the agent loop, tools, and model-facing surfaces live in the SDK.
  \item[Typed loop I/O.] Limited. The loop boundary is natural language: \texttt{Conversation.send\_message} accepts \texttt{str | Message} (user text/image content) and the run's result is a string -- \texttt{get\_agent\_final\_response(events)} returns the text of \texttt{FinishAction.message} (or the agent's last message). Tool calls are validated into typed pydantic \texttt{Action}/\texttt{Observation} models (\texttt{ToolDefinition[ActionT, ObservationT]}), but there is no typed task input and no structured-output path (\texttt{response\_format} appears nowhere in the SDK).
  \item[Pass by reference.] Partial. Data passes between steps through workspace files: \\ \texttt{FileEditorAction.path} takes an absolute path, and sub-agent tasks and results are plain text. The only live handles the model can name are session identifiers -- \texttt{TaskAction.resume} takes a task ID to resume a sub-agent -- not data references. File-based passing caps the axis at Partial.
  \item[Code as action.] Partial. The CodeAct-era IPython cell is gone: the V1 SDK's default preset is discrete tools (\texttt{TerminalTool}, \texttt{FileEditorTool}, \texttt{TaskTrackerTool}, browser), and \texttt{TerminalAction.command} executes one shell command in a persistent PTY terminal, with \texttt{is\_input} for sending keystrokes to a running process -- a documented pattern for driving interactive programs -- including interactive interpreters -- inside the shell tool; no general-purpose Python execution tool exists in either pinned tree. A \texttt{WorkflowTool} does let the model write Python with control flow and inline agent calls (\texttt{wf.run\_agent}, \texttt{wf.map\_agents}, \texttt{wf.pipeline}), but it is absent from every preset, barred from direct file and shell work, and self-described as an MVP\textsuperscript{\dag}.
  \item[Loop engineering.] Strong (opt-in). Developers program the loop directly: \texttt{AgentBase.step} is abstract, \texttt{Conversation.run} is a pausable while-loop, and condensers, hooks, and a \texttt{GoalController} compose around it. The model orchestrates via a delegation tool\textsuperscript{\dag}: \texttt{TaskToolSet} spawns registered sub-agent types (including file-based \texttt{.md} agent definitions) and resumes them by task ID, behind the \texttt{enable\_sub\_agents} setting (default off); a spawn/delegate executor also exists in the tree as unwired scaffolding with no registered tool. The \texttt{WorkflowTool} lets the model author orchestration loops in Python (\texttt{wf.run\_agent}, \texttt{wf.map\_agents}, \texttt{wf.pipeline}); it is self-described as an MVP and in no preset\textsuperscript{\dag}.
  \item[Object state.] Partial. The default preset ships \texttt{TaskTrackerTool}, a read/write typed todo store: \texttt{view}/\texttt{plan} commands over a \texttt{list[TaskItem]} (title, notes, status), persisted to \texttt{TASKS.json}. The planning preset keeps its state in a \texttt{PLAN.md} file via \texttt{PlanningFileEditorTool}, and \texttt{ThinkTool} is a no-op scratchpad. Todo tools and plan files cap the axis at Partial; there is no typed, live object store.
  \item[Harness APIs.] Partial. Skill loading is model-callable: the \texttt{invoke\_skill} builtin is ``the only supported way to invoke a skill'' listed in \texttt{<available\_skills>} and is auto-attached whenever an invocable skill is loaded. Everything else is harness- or dev-side: keyword/task/path triggers auto-inject skill content into user messages, condensation is a developer/user API (\texttt{conversation.condense()}; never model-callable), and hooks are dev lifecycle scripts. Other model-callable controls exist -- \texttt{switch\_llm} (default-on LLM-profile switching) and \texttt{vision\_inspect} (auto-attached for models without vision)\textsuperscript{\dag} -- but neither is a context or event API; loading alone caps the axis at Partial.
\end{description}

\subsection{PI}
\begin{description}
  \item[Evidence base.] \texttt{earendil-works/pi} (formerly \texttt{badlogic/pi-mono}) at commit \\ \texttt{351efc828b6fc5250fa50d6b32b20b0f0cb22cb4} (coding-agent 0.80.3). The benchmark comparisons elsewhere in this paper use PI v0.72.1; re-verifying the capability scores at 0.80.3 produced no changes.
  \item[Typed loop I/O.] Limited. Tool inputs are TypeBox-schema typed, but tool results return to the model as untyped content blocks (typed \texttt{details} never reach the model) and agent-level I/O is prompt-in/text-out; a typed final output exists only as an opt-in example extension.
  \item[Pass-by-reference.] Partial. The model works through files and shell output \\(\texttt{read}/\texttt{bash}/\texttt{edit}/\texttt{write}); extensions can hold live in-process state shared across the tools the model calls, but the model itself handles only paths and text.
  \item[Code as action.] Partial. The open-ended action modality is the \texttt{bash} tool; there is no code cell with inline harness method/tool calls.
  \item[Loop engineering.] Partial. Developers orchestrate via the TypeScript SDK (\texttt{createAgentSession}) and extensions; a shipped opt-in example extension gives the model a subagent tool with single/parallel/chain modes, but the model fills structured parameters rather than authoring loops, and core PI deliberately ships without subagents.
  \item[Object state.] Limited. Session trees, forks, and extension persistence (\texttt{appendEntry}) are user/developer-facing; no durable state store is model-visible by default (a stateful todo tool exists only as an example extension; skills listings and \texttt{AGENTS.md} files are injected as prompt text).
  \item[Harness APIs.] Partial. Skill loading is first-class, default-on, and model-initiated: skills are auto-discovered from user and project directories, advertised in the system prompt, and the prompt instructs the model to load a skill's file with the \texttt{read} tool (a per-skill \texttt{disable-model-invocation} flag exists precisely to turn this off). Loading only; the extension context/event APIs (\texttt{before\_agent\_start}, \texttt{context} events, event bus) remain developer-only.
\end{description}

\subsection{Hermes}
\begin{description}
  \item[Evidence base.] \texttt{NousResearch/hermes-agent} at commit \\\texttt{3c63ed3a3c81fd3d924128f4be51df7e7c21cd06} (hermes-agent 0.18.0).
  \item[Typed loop I/O.] Limited. The embedding API is \texttt{agent.chat(prompt) $\rightarrow$ str} with no structured-output machinery at the loop boundary. Tool inputs are JSON-schema typed; tool and subagent returns are text or JSON strings (vision and browser tools can add image blocks).
  \item[Pass-by-reference.] Partial. Oversized tool outputs are spilled to files and replaced in-context by a preview plus a dereferenceable path the model can \texttt{read\_file} and page through; the same pattern covers oversized subagent summaries. There are no live object references.
  \item[Code as action.] Strong. \texttt{execute\_code} lets the model write a Python script whose inline RPC calls (\texttt{web\_search}, \texttt{web\_extract}, \texttt{read\_file}, \texttt{write\_file}, \texttt{search\_files}, \texttt{patch}, foreground \texttt{terminal}) are real function calls returning parsed values; intermediate tool results never enter the context window -- only printed output (plus stderr on failure) comes back.
  \item[Loop engineering.] Partial. \texttt{delegate\_task} supports parallel batch fan-out and nested orchestrator roles bounded by spawn depth (config-gated; flat by default) (top-level delegations always run in the background; blocking mode is depth-determined, not model-chosen); but model-authored orchestration code is blocked by design -- \texttt{execute\_code} and \texttt{delegate\_task} are mutually excluded from each other's reach.
  \item[Object state.] Partial. The default toolset includes a \texttt{memory} tool (add/replace/remove) over \texttt{MEMORY.md}/\texttt{USER.md}, injected into the system prompt as a snapshot with capacity metadata, plus \texttt{skill\_manage} for durable procedural skills. Entries are bounded text; the prompt snapshot is frozen per session, with live state visible only through the tool -- not live fields.
  \item[Harness APIs.] Partial. Model-callable surfaces include the \texttt{memory} block, \texttt{session\_search} (full-text search over past session transcripts), and \texttt{tool\_search}/\texttt{tool\_call} progressive tool disclosure (auto-enabled only when deferrable MCP/plugin tool schemas exceed a context threshold). Per-turn context assembly and compaction remain automatic and developer-facing, and there is no live event-stream API.
\end{description}

\subsection{OpenCode}
\begin{description}
  \item[Evidence base.] \texttt{sst/opencode} (\texttt{dev} branch) at commit \texttt{1c25b2f298d49d89ce473646de5766aa754c59f2} (opencode 1.17.14).
  \item[Typed loop I/O.] Limited. Tool inputs are schema-typed and validated, but every tool and subagent return is a plain string (task results come back XML-wrapped); there is no typed loop contract. (Inside the experimental CodeMode cell, tool calls do carry both input and output schemas.)
  \item[Pass-by-reference.] Partial. The model works through file paths, diffs, and text observations; an \texttt{experimentalReferences} flag exists but is unused. Within an experimental CodeMode cell, one tool's structured result can flow to the next call as an interpreter value, but it is JSON-only and ephemeral per cell.
  \item[Code as action.] Strong (flag-gated). The flag-gated experimental CodeMode \texttt{execute} tool is a genuine code cell with inline schema-typed tool calls; its catalog covers MCP tools only and it is off by default\textsuperscript{\dag}. The default modality is shell plus file edits.
  \item[Loop engineering.] Partial. The model delegates via the \texttt{task} tool (one subagent per call, resumable, with a separately flag-gated background option); developers orchestrate through the HTTP server, JS SDK, and plugin hooks. There are no model-authored agent loops.
  \item[Object state.] Limited. \texttt{todowrite} persists a typed todo list to SQLite beyond the transcript, but there is no read-back tool and todos are not re-injected into context -- the model sees its state only through its own past tool outputs. Sessions, snapshots, and revert are user/developer checkpoints.
  \item[Harness APIs.] Partial. The \texttt{skill} tool is model-callable and on by default: available skills are listed in the system prompt and the model loads a skill's instructions and file manifest on demand. Loading only -- plugin hooks can transform system prompts and messages per turn and there is a developer event bus, but those remain hidden developer callbacks, and nothing is model-callable for context blocks or events.
\end{description}

\subsection{OpenClaw}
\begin{description}
  \item[Evidence base.] \texttt{openclaw/openclaw} at commit \texttt{29f787f10ed4539c410749ef33a2d64928c9be0f} (version 2026.6.10). OpenClaw describes itself as a personal AI assistant and multi-channel gateway with an embedded agent runtime.
  \item[Typed loop I/O.] Limited. Tool inputs are TypeBox-typed but tool returns are text (or image) blocks, plain or JSON-stringified, and agent-to-agent invocation is free text (\texttt{sessions\_spawn} takes a task string; results return as announced text). An \texttt{outputSchema} field exists only as unused metadata.
  \item[Pass-by-reference.] Partial. Same-agent subagent spawns inherit the parent workspace, so files are stable shared referents across those sessions; everything else crosses serialized gateway boundaries, and spawn attachments are explicitly snapshot-by-value.
  \item[Code as action.] Strong (opt-in). An opt-in Code Mode cell (QuickJS) gives the model inline \texttt{tools.call}/namespace calls with object results; when enabled it replaces the normal tool surface (only the cell's \texttt{exec}/\texttt{wait} are exposed), and it is off by default\textsuperscript{\dag}. The default modality is the \texttt{exec} shell tool under approval/policy controls alongside read/write/edit tools.
  \item[Loop engineering.] Strong (opt-in). Session tools (\texttt{sessions\_list}, \texttt{sessions\_history}, \\ \texttt{sessions\_send}, \texttt{sessions\_spawn}, \texttt{sessions\_yield}, \texttt{subagents}) give the model event-driven, tool-mediated orchestration -- spawning one-shot or persistent children, messaging them, and yielding until completion events (polling loops are explicitly discouraged). With Code Mode enabled the session tools are in the cell's catalog, so model-written JavaScript can author spawn loops\textsuperscript{\dag} -- though child results arrive as events after the turn, not as in-cell values.
  \item[Object state.] Partial. The model durably writes and re-reads its own state: \texttt{MEMORY.md} (re-injected at session bootstrap) and \texttt{memory/*.md} notes, written via ordinary file tools and queried through model-callable \texttt{memory\_search}/\texttt{memory\_get}; \texttt{session\_status} also lets the model update the session's model override. These are files and tool-mediated state, not live object fields.
  \item[Harness APIs.] Partial. \texttt{session\_status}, \texttt{sessions\_history} (sanitized transcript inspection with pagination), and the memory tools are model-callable; the context-assembly mechanism (which files are compiled into the system prompt, budgets, compaction) remains host/developer controlled, though the model can author the bootstrap files' contents.
\end{description}

\subsection{\nooa}
\begin{description}
  \item[Evidence base.] \nooa source at commit \texttt{59ae56bf47af7582990355ad5312da098713c8f4} (2026-07-01).
  \item[Typed loop I/O.] Strong. Generation-method arguments are validated on every call, and the default CodeAct strategy (like Predict) requires a return annotation and validates returned values -- with validation errors fed back to the model for retry. Deterministic (tool) methods called from generated code are ordinary Python calls; their values are validated against the generation method's annotations when they cross a generation boundary.
  \item[Pass-by-reference.] Strong. Live Python values are model-visible by name, type, and bounded preview, and can be passed between methods/tools without serializing the full object into the prompt. This is a property of the code-executing strategies (CodeAct, the default, and PurePython); \texttt{PredictStrategy} instead serializes full argument representations into the prompt, guarded by a hard size cap that fails loudly rather than truncating.
  \item[Code as action.] Strong. CodeAct cells are Python actions with control flow, helper functions, and inline method or tool calls, returning Python values or stdout as observations; a safety validator screens cells for hazards (\texttt{eval}/\texttt{exec}, blocking calls, unbounded loops) before execution.
  \item[Loop engineering.] Strong. Developers and models use ordinary Python to create subagents, call strategies, and write produce/evaluate/retry orchestration loops over the same callable primitives; the CodeAct system prompt itself documents the fan-out pattern (\texttt{@strategy}-decorated standalone functions run in parallel with \texttt{asyncio.gather}).
  \item[Object state.] Strong. State lives on \texttt{self} and ordinary Python objects, rendered into a bounded per-turn state block that is re-evaluated on every LLM turn, so the agent can store, inspect, and reuse explicit state without reconstructing it from a message transcript. Data fields only: attaching method-like callables to public attributes of \texttt{self} is blocked by design, and helpers live in REPL scope.
  \item[Harness APIs.] Strong. Structured context blocks, per-turn dynamic context such as skill/tool inventory blocks, and queryable session events are first-class runtime APIs shared by developers and the model. As a context-hygiene default, \texttt{self.context} and \texttt{self.events} are omitted from the agent's self-documentation until the developer opts them into visibility per instance; once exposed they are fully model-callable.
\end{description}

\input{appendix_stress_trace}

\input{appendix_memory_arc3}

\end{document}

%% file: codeact_loop_v2.tex
\begin{figure}[t]
\centering
\begin{tikzpicture}[
  >={Triangle[]}, font=\footnotesize, node distance=9mm and 8mm,
  stage/.style={draw=builtinblue!65, line width=0.8pt, rounded corners=4pt,
                top color=white, bottom color=builtinblue!8,
                align=center, text width=0.165\linewidth, minimum height=1.25cm,
                inner sep=4pt},
  endpoint/.style={draw=nvgreen!55!black, line width=0.9pt, rounded corners=9pt,
                   top color=white, bottom color=nvgreen!14,
                   align=center, text width=0.20\linewidth, minimum height=0.95cm,
                   inner sep=4pt},
  badge/.style={circle, fill=builtinblue, text=white, font=\scriptsize\bfseries,
                inner sep=0pt, minimum size=4.4mm, draw=white, line width=0.7pt},
  flow/.style={-{Triangle[length=2.6mm,width=2.4mm]}, line width=1pt,
               builtinblue!72!black},
  loop/.style={-{Triangle[length=2.6mm,width=2.4mm]}, line width=1pt,
               nvgreen!48!black},
  pill/.style={fill=nvgreen!10, draw=nvgreen!55!black, line width=0.5pt,
               rounded corners=3pt, inner xsep=5pt, inner ysep=2.5pt,
               font=\scriptsize\itshape},
]
  \node[stage] (render) {
    \textbf{Render Context}\\[-1pt]
    {\scriptsize Sec.~\ref{sec:agent-loop-render-context}}
  };

  \node[stage, right=of render] (llm) {
    \textbf{Call LLM}\\[-1pt]
    {\scriptsize Sec.~\ref{sec:agent-loop-call-llm}}
  };

  \node[stage, right=of llm] (exec) {
    \textbf{Execute Python}\\[-1pt]
    {\scriptsize Sec.~\ref{sec:agent-loop-execute-python}}
  };

  \node[stage, right=of exec] (update) {
    \textbf{Update State}\\[-1pt]
    {\scriptsize Sec.~\ref{sec:agent-loop-update-events-state}}
  };

  \node[endpoint, above=14mm of render] (entry) {
    \textbf{Call}\\[-1pt]
    %{\scriptsize Sec.~\ref{sec:agentic-methods}}
  };

  \node[endpoint, below=11mm of update] (out) {
    \textbf{Return}\\[-1pt]
    {\scriptsize Sec.~\ref{sec:agent-loop-strong-typing}}
  };

  \draw[flow] (entry) -- (render);
  \draw[flow] (render) -- (llm);
  \draw[flow] (llm) -- (exec);
  \draw[flow] (exec) -- (update);

  \draw[loop, rounded corners=7pt]
        (update.east)
        -- ([xshift=8mm]update.east)
        -- ([xshift=8mm,yshift=8mm]update.north east)
        -- ([xshift=-9mm,yshift=8mm]render.north west)
        -- ([xshift=-9mm]render.west) -- (render.west);

  \node[pill] at ([yshift=8mm]$(render.north)!0.5!(update.north)$)
        {\textbf{Next Turn:} updated events \& state};

  \draw[loop] (update.south) -- node[pill, pos=0.4, right=1.5mm]{validated} (out.north);

  \begin{scope}[on background layer]
    \node[draw=codeframe, line width=1pt, rounded corners=7pt, fill=black!2,
          fit=(render)(llm)(exec)(update), inner sep=9pt] (turnbox) {};
  \end{scope}
\end{tikzpicture}
\caption{\textbf{The CodeAct strategy loop within an agentic method.}
A caller invokes the method, then each turn renders context, calls the LLM, executes Python actions,
and updates events and state. Once a successful, type-validated value is recorded, it is returned to
the caller.}
\label{fig:agent-loop}
\end{figure}

%% file: memory_system.tex
% ===========================================================================
\subsection{Long-Term Memory: the Agent Curates Its Own State}
\label{sec:memory}
% ===========================================================================
The mechanisms described so far are scoped to a method call or a session, yet an agent with frozen weights can only improve through the state it retains. Our companion work on workspace optimization~\cite{sarafian2026workspace} shows that agents can learn by writing typed, evidence-gated artifacts in place of parameter updates; its principal open problem is \emph{transfer}, because the workspace is discarded at every task boundary. \nooa addresses transfer with an optional long-term memory subsystem: \inlinecode{\texttt{MemoryManager.install(agent)}} attaches it to an unmodified agent, and uninstalling restores the agent exactly.

\inlineparagraph{The agent authors its own memory} Following Principle~5, writing a memory is a deliberate action of the model rather than the output of a background extraction pipeline. Seven model-callable tools (\inlinecode{\texttt{remember}}, \inlinecode{\texttt{recall}}, \inlinecode{\texttt{search}}, \inlinecode{\texttt{update\_memory}}, \inlinecode{\texttt{forget}}, \inlinecode{\texttt{associate}}, \inlinecode{\texttt{deref}}) operate on the store; they accept ordered verbal descriptors (\textsc{critical}\,\ldots\,\textsc{trivial}) that map to numeric scores internally, and a standing context block states that the store is the agent's own to maintain.

\inlineparagraph{Deliberate and spontaneous recall} Memory reaches the model through two channels: the agent queries the store with its tools, and a \inlinecode{\texttt{BeforeTurn}} hook derives a query from recent events and injects associated memories into a dynamic context block. Injected memories are not reinforced, so what the harness surfaces does not distort the usage signal. Retrieval unions embedding and keyword candidates, ranks them by ACT-R activation~\cite{anderson2004act} -- relevance, recency, and importance, the triad of generative agents~\cite{park2023generative} -- and propagates activation over a typed memory graph. Decay-based forgetting keeps the store bounded.

\inlineparagraph{Asynchronous reflection} Consolidation runs outside the agent loop, after a task completes or while the agent is idle, as an ordered pass: near-duplicate memories are merged; conflicting values can be reconciled into a single current record, archiving the superseded ones; related memories are linked; importance is re-scored; episodes can be distilled into higher-level records; and memories whose activation has decayed are pruned. Pruning never removes recent memories, protected types, or open todos.

\begin{figure}[tbp]
\centering
\includegraphics[width=0.96\linewidth]{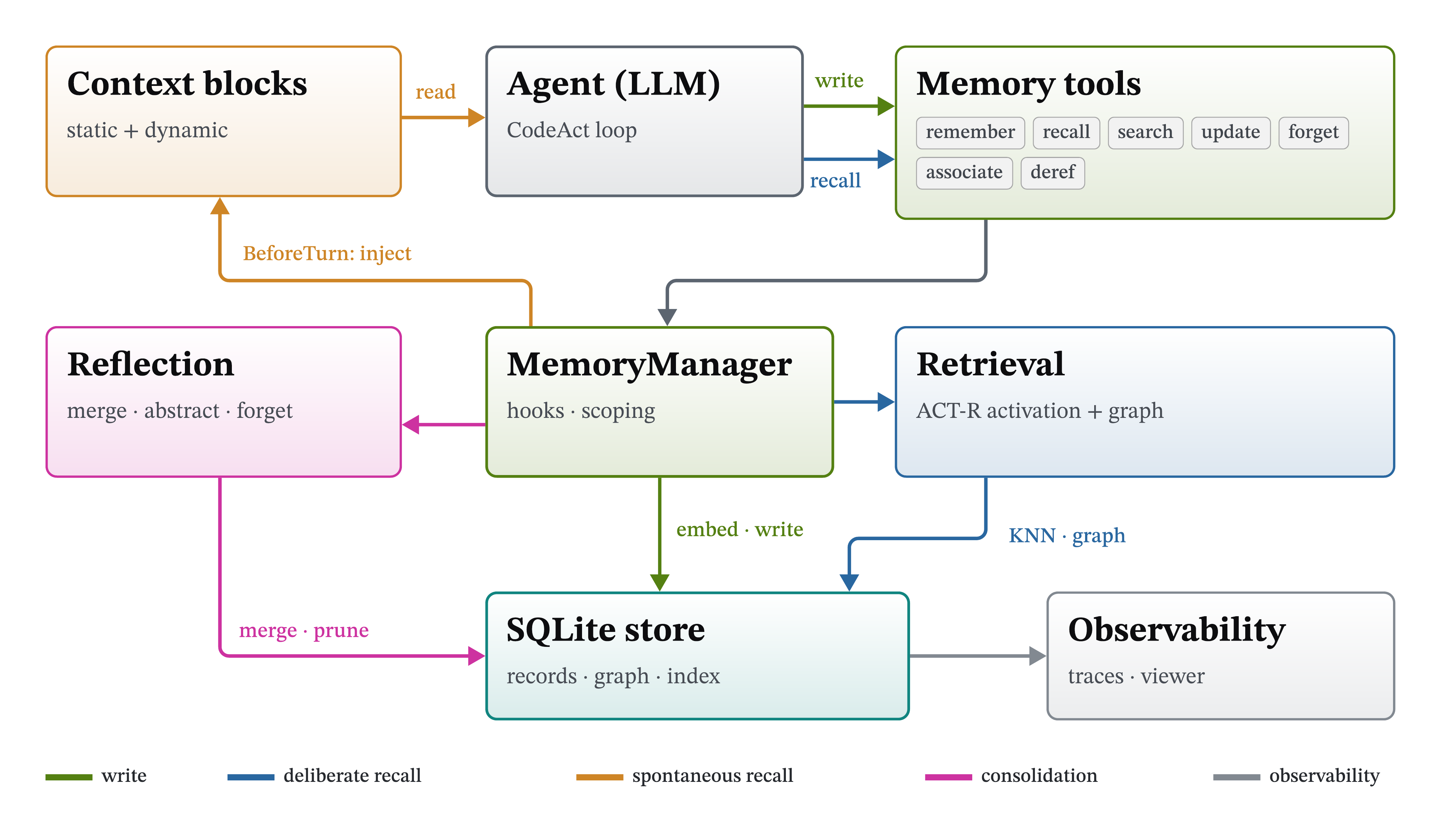}
\caption{\textbf{The \nooa memory system.} \inlinecode{\texttt{MemoryManager.install(agent)}} attaches memory to an unmodified agent. The agent curates its own store through seven tools (write, green; deliberate recall, blue); a \inlinecode{\texttt{BeforeTurn}} hook injects associated memories into a dynamic context block (spontaneous recall, amber); reflection consolidates the store (magenta); every access is recorded (gray). All state lives in one human-inspectable SQLite file; the vector index is derived and pluggable.}
\label{fig:memory-arch}
\end{figure}

\inlineparagraph{One inspectable file; live references} The entire store is one SQLite file that can be inspected directly; vector indexes are derived from it and interchangeable. A memory may hold typed references (\inlinecode{\texttt{kind:key}}) that are resolved against live agent state at recall time -- extending pass by reference into persistence, so recall does not answer from stale copies -- and owner scoping governs reads and writes when several agents share one store. The subsystem's end-to-end effect is measured in Sec.~\ref{sec:arc3}: +11.8 RHAE points over the identical agent with file-based notes in place of memory. Figure~\ref{fig:memory-arch} shows the architecture; Appendix~\ref{app:memory-details} details the design and compares memory support across contemporary harnesses.

%% file: swe_results_v2.tex
SWE-bench Verified~\cite{jimenez2024swebench} contains 500 software-engineering tasks derived from issues in real GitHub repositories. An agent must inspect an unfamiliar codebase, identify the cause of a reported problem, modify the repository, and produce a patch that passes the benchmark's tests. Terminal-Bench~2.0~\cite{tbench_2025} contains 89 tasks performed through a command-line environment, including software installation, configuration, debugging, and service operation.

\inlineparagraph{Agent and comparison harnesses}
For both benchmarks, we use the same benchmark-agnostic agent, \inlinecode{\texttt{BenchAgent}}. The agent has a todo list, shell tools for command execution and file editing, and repository-navigation tools based on tree-sitter. Its dynamic context contains the task description, todo-list status, context-window statistics, and the current working state of its shell and repository tools. The agent terminates through a typed \inlinecode{\texttt{TaskResult}} containing the identified root cause, supporting evidence, and a verification command. This return value is validated by the harness before execution ends. The complete agent consists of 253 lines of ordinary Python and is included in the \nooa repository.

\begin{figure}[t]
\caption{\textbf{SWE-Bench Verified score vs.\ per-task prefill+output token cost.} Color encodes
harness, marker shape encodes backend family (circle = GPT 5.5, square = Opus 4.6), and marker size
encodes reasoning effort (off\,$<$\,high\,$<$\,xhigh).}
\label{fig:pareto}
\definecolor{cNOOA}{RGB}{31,119,180}    % blue
\definecolor{cBASE}{RGB}{23,165,137}    % teal (unused; NOOA-base removed)
\definecolor{cOC}{RGB}{230,126,34}      % orange
\definecolor{cPI}{RGB}{155,89,182}      % purple

% marker size = reasoning effort
\newcommand{\szoff}{1.7}
\newcommand{\szhi}{3.2}
\newcommand{\szxhi}{4.8}
\newcommand{\sqoff}{1.7}
\newcommand{\sqhi}{3.2}

\tikzset{
  lead/.style={gray!55, line width=0.3pt},
  lab/.style={font=\scriptsize, inner sep=1pt},
}
\centering
\begin{tikzpicture}
\begin{semilogxaxis}[
  width=15cm, height=8.4cm,
  xlabel={Mean tokens per task (prefill + output, $\times 10^3$, log scale)},
  ylabel={SWE-bench Verified pass rate (\%)},
  xmin=80, xmax=3200, ymin=56, ymax=88,
  grid=major, grid style={black!10},
  legend cell align=left,
  legend style={
    at={(0.5,-0.15)}, anchor=north, legend columns=4,
    font=\footnotesize, draw=black!30,
    /tikz/every even column/.append style={column sep=8pt},
  },
  tick label style={font=\footnotesize}, label style={font=\small},
]

% ===== Pareto frontier (gpt + opus, without NOOA-base) =====
\addplot[black!55, dashed, thick, mark=none] coordinates {
  (103,60.8) (276,67.2) (588,75.8) (842,79.8) (1075,82.2)};
\addlegendentry{Pareto frontier}

% ===== leader lines =====
\draw[lead] (axis cs:791,75.2) -- (axis cs:770,67);
\draw[lead] (axis cs:842,79.8) -- (axis cs:800,86);
\draw[lead] (axis cs:843,78.8) -- (axis cs:980,77);
\draw[lead] (axis cs:858,75.0) -- (axis cs:910,66);
\draw[lead] (axis cs:1075,82.2) -- (axis cs:1075,85.7);
\draw[lead] (axis cs:1277,78.6) -- (axis cs:1330,81.7);
\draw[lead] (axis cs:1341,73.6) -- (axis cs:1341,70.5);

% ===== markers: color=harness, shape=backend, size=effort =====
% NOOA (blue)
\addplot[forget plot,only marks,mark=*,      mark size=\szoff pt, cNOOA] coordinates {(276,67.2)};
\addplot[forget plot,only marks,mark=*,      mark size=\szhi pt,  cNOOA] coordinates {(843,78.8)};
\addplot[forget plot,only marks,mark=*,      mark size=\szxhi pt, cNOOA] coordinates {(1075,82.2)};
\addplot[forget plot,only marks,mark=square*,mark size=\sqhi pt,  cNOOA] coordinates {(842,79.8)};
% OpenCode (orange)
\addplot[forget plot,only marks,mark=*,      mark size=\szoff pt, cOC] coordinates {(217,59.2)};
\addplot[forget plot,only marks,mark=*,      mark size=\szhi pt,  cOC] coordinates {(858,75.0)};
\addplot[forget plot,only marks,mark=*,      mark size=\szxhi pt, cOC] coordinates {(1277,78.6)};
\addplot[forget plot,only marks,mark=square*,mark size=\sqhi pt,  cOC] coordinates {(791,75.2)};
% PI (purple)
\addplot[forget plot,only marks,mark=*,      mark size=\szoff pt, cPI] coordinates {(103,60.8)};
\addplot[forget plot,only marks,mark=*,      mark size=\szhi pt,  cPI] coordinates {(1341,73.6)};
\addplot[forget plot,only marks,mark=*,      mark size=\szxhi pt, cPI] coordinates {(2212,78.2)};
\addplot[forget plot,only marks,mark=square*,mark size=\sqhi pt,  cPI] coordinates {(588,75.8)};

% ===== legend: harness colors only =====
\addlegendimage{cNOOA, line width=3.2pt, mark=none}\addlegendentry{NOOA}
\addlegendimage{cOC,   line width=3.2pt, mark=none}\addlegendentry{OpenCode}
\addlegendimage{cPI,   line width=3.2pt, mark=none}\addlegendentry{PI}

% ===== labels (backend-effort), color = harness =====
% gpt-off (left)
\node[lab,cPI,  anchor=north] at (axis cs:103,60.0) {gpt-off};
\node[lab,cOC,  anchor=south] at (axis cs:217,60.3) {gpt-off};
\node[lab,cNOOA,anchor=west]  at (axis cs:305,66.8) {gpt-off};
% opus-high
\node[lab,cPI,  anchor=east]  at (axis cs:573,75.8) {opus-high};
\node[lab,cOC,  anchor=north] at (axis cs:770,66.4) {opus-high};
\node[lab,cNOOA,anchor=south] at (axis cs:750,86.0) {opus-high};
% gpt-high
\node[lab,cNOOA,anchor=west]  at (axis cs:985,76.0) {gpt-high};
\node[lab,cOC,  anchor=north] at (axis cs:910,65.4) {gpt-high};
\node[lab,cPI,  anchor=north] at (axis cs:1341,70.3) {gpt-high};
% gpt-xhigh
\node[lab,cNOOA,anchor=south] at (axis cs:1075,85.7) {gpt-xhigh};
\node[lab,cOC,  anchor=south] at (axis cs:1370,80.7) {gpt-xhigh};
\node[lab,cPI,  anchor=south] at (axis cs:2212,79.6) {gpt-xhigh};

\end{semilogxaxis}
\end{tikzpicture}
\end{figure}
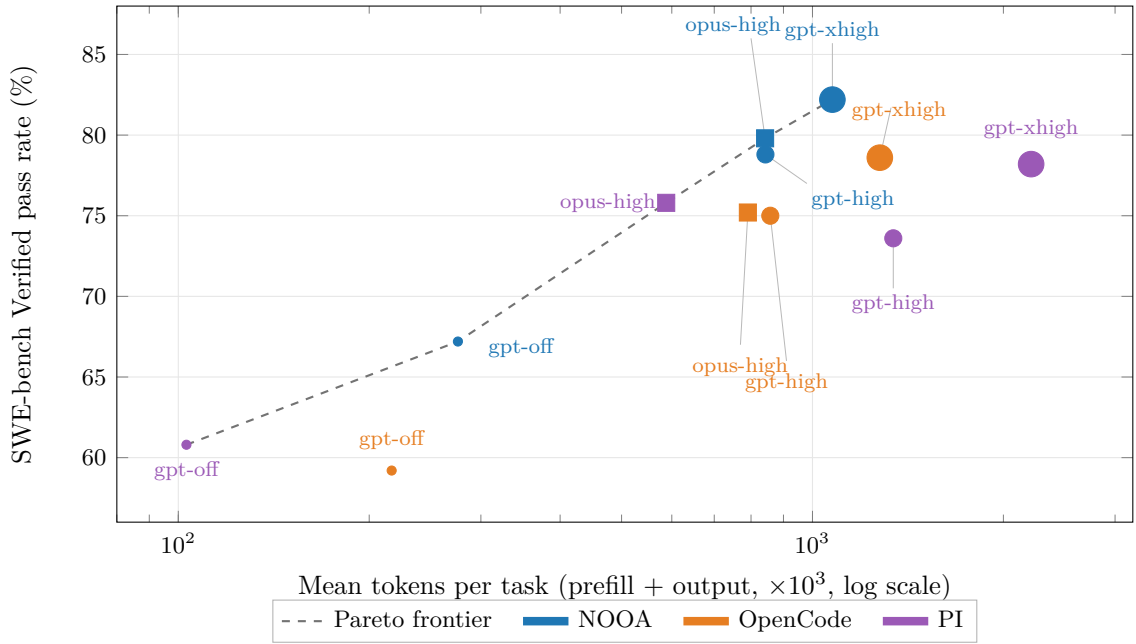

We compare \nooa with two open, general-purpose coding agents. OpenCode~\cite{opencode} is a full-featured terminal coding agent with file, search, and shell tools, together with automatic transcript summarization. PI~\cite{pi} is a deliberately minimal agent with a small prompt, standard file and shell tools. All three harnesses are evaluated with the same GPT-5.5 and Claude Opus~4.6 backends at the available reasoning-effort settings. We report task pass rate in Tables~\ref{tab:swe} and~\ref{tab:tb2}, and per-task token usage in Figure~\ref{fig:pareto}.

\inlineparagraph{Results}
On SWE-bench Verified, \nooa obtains the highest pass rate among the open harnesses in every evaluated model and reasoning configuration. 
Results show that NOOA improves on the original CodeAct paradigm as implemented by OpenHands v3~\cite{openhands2024}, which under Opus 4.6 is reported to have a 68.4\% pass rate. \nooa builds on this result by 11.4 points, given the same model. With GPT-5.5, it reaches 67.2\%, 78.8\%, and 82.2\% at off, high, and xhigh reasoning effort, respectively. At xhigh effort, OpenCode reaches 78.6\% and PI reaches 78.2\%. With Opus~4.6, \nooa reaches 79.8\%, compared with 75.2\% for OpenCode and 75.8\% for PI.

The advantage is larger on Terminal-Bench~2.0. With GPT-5.5 and reasoning disabled, \nooa reaches 46.1\%, compared with 34.8\% for OpenCode and 37.1\% for PI. At high effort, it reaches 73.0\%, ahead of OpenCode by 12.3 points and PI by 4.5 points. PI obtains the best GPT-5.5 xhigh result at 75.3\%, compared with 73.0\% for \nooa. With Opus~4.6 at high effort, \nooa reaches 65.2\%, while OpenCode and PI reach 43.8\% and 58.4\%, respectively.

The higher pass rates do not come from using longer trajectories. On SWE-bench with GPT-5.5 xhigh, \nooa reaches 82.2\% using approximately 28 model calls and 1.1 million tokens per task. OpenCode uses a similar number of calls but approximately 1.3 million tokens for 78.6\%, while PI uses 66 calls and 2.2 million tokens for 78.2\%. As shown in Figure~\ref{fig:pareto}, \nooa therefore defines most of the observed accuracy--cost frontier.

\paragraph{Effect of reasoning effort.}
Increasing reasoning effort improves all three harnesses, but the interface matters most when the model provides less planning and verification discipline of its own. With reasoning disabled, \nooa leads OpenCode and PI by 8.0 and 6.4 points on SWE-bench, and by 11.3 and 9.0 points on Terminal-Bench. These margins narrow at higher effort, suggesting that the explicit object state, typed actions, and programmable loop behavior exposed by \nooa partly substitute for behaviors that stronger reasoning models increasingly perform themselves.

\paragraph{Validated termination.}
Trace analysis identifies termination as a key difference between the harnesses. OpenCode stops whenever the model responds without a tool call; on Terminal-Bench, 77\% of its failed GPT-5.5 trials terminate within ten steps. In \nooa, the model must instead return a validated \inlinecode{\texttt{TaskResult}} containing evidence and a verification command. This prevents unsupported declarations of completion and is especially valuable on tasks whose intermediate state can appear correct before hidden checks are run. More broadly, it illustrates the value of treating type annotations as executable contracts: termination becomes a programmatically validated action rather than an informal convention encoded only in the prompt.

\paragraph{Interaction and context efficiency.}
\nooa also uses fewer tokens because tool outputs remain available as live Python values rather than being repeatedly serialized through the transcript. With GPT-5.5 xhigh on SWE-bench, it reaches 82.2\% using approximately 28 calls and 1.1M tokens per task, compared with 78.2\% using 66 calls and 2.2M tokens for PI. Bounded prompt previews also keep \nooa well below the context limit, avoiding the lossy transcript compaction used by OpenCode and PI while preserving prefix-cache reuse. These results directly expose the benefits of combining code as action with pass-by-reference: the model can operate on persistent objects in the execution environment instead of repeatedly exchanging their full textual representations with the harness.

\paragraph{Comparison with specialized systems.}
The results also narrow the gap between open general-purpose harnesses and specialized closed systems. On SWE-bench Verified, \nooa reaches 82.2\% with GPT-5.5 and 79.8\% with Opus~4.6, compared with 88.7\% for Codex and 80.8\% for Claude Code. On Terminal-Bench~2.0, its 65.2\% with Opus~4.6 is comparable to the 62.9--65.4\% reported for Claude Code and Terminus-2. Thus, a small benchmark-agnostic \nooa agent is competitive with specialized systems while consistently outperforming the open general-purpose harnesses in our comparison. This supports the broader agent-as-a-Python-object claim: ordinary classes, methods, state, and type contracts provide a simple developer-facing abstraction without making the interface less effective for models.

\begin{table}[t]
\centering
\setlength{\tabcolsep}{3pt}
%---------------------------- left: SWE-bench ----------------------------
\begin{minipage}[t]{0.46\linewidth}
\centering
\captionof{table}{SWE-bench Verified pass rates. Published leaderboard SOTA at submission: 79.2\% with a specialized agent + Opus 4.5.}
\label{tab:swe}
\small
\scalebox{0.96}{
\begin{tabular}{l ccc cc}
\toprule
& \multicolumn{3}{c}{GPT-5.5} & \multicolumn{2}{c}{Opus 4.6} \\
\cmidrule(lr){2-4}\cmidrule(lr){5-6}
Harness & off & high & xhigh & off & high \\
\midrule
% NOOA-base & 60.2 & \textbf{80.2} & \textbf{84.0} & 76.2 & 75.6 \\
NOOA      & \textbf{67.2} & \textbf{78.8} & \textbf{82.2} & \textbf{76.8} & \textbf{79.8} \\
OpenCode 1.14.33 & 59.2 & 75.0 & 78.6 & 76.0 & 75.2 \\
PI v0.72.1 & 60.8 & 73.6 & 78.2 & 75.6 & 75.8 \\
\bottomrule
\end{tabular}
}
\end{minipage}\hfill
%---------------------------- right: Terminal Bench 2.0 ----------------------------
\begin{minipage}[t]{0.46\linewidth}
\centering
\captionof{table}{Terminal Bench 2.0 (89 tasks), task pass rate (\%). Published leaderboard SOTA at submission: 84.7\% (NexAU-AHE + GPT-5.5).}
\label{tab:tb2}
\small
\scalebox{0.96}{
\begin{tabular}{l ccc cc}
\toprule
& \multicolumn{3}{c}{GPT-5.5} & \multicolumn{2}{c}{Opus 4.6} \\
\cmidrule(lr){2-4}\cmidrule(lr){5-6}
Harness & off & high & xhigh & off & high \\
\midrule
% NOOA-base & 36.0 & \textbf{77.5} & 71.9 & \textbf{67.4} & 61.8 \\
NOOA      & \textbf{46.1} & \textbf{73.0} & 73.0 & 64.0 & \textbf{65.2} \\
OpenCode 1.14.33 & 34.8 & 60.7 & 52.8 & 49.4 & 43.8 \\
PI v0.72.1       & 37.1 & 68.5 & \textbf{75.3} & \textbf{65.2} & 58.4 \\
\bottomrule
\end{tabular}}
\end{minipage}
\end{table}

%% file: cybergym.tex
CyberGym~\cite{wang2026cybergym} is a security benchmark, in which an agent must inspect a codebase, identify a security-relevant bug, and validate it by producing a proof-of-concept that reliably triggers it.
Agentic vulnerability discovery is notoriously difficult and can consume large amounts of context.
Crash reports are long; code bases can be long; and small pieces of information need to be coupled across potentially long distances.
We test whether the deconstruction and simplification offered by the \nooa architecture can yield gains in the vulnerability validation stage.

\inlineparagraph{CyberGym \nooa agent} Runs in the trial container as a CodeAct agent with shell and a todo manager tools. The agent reads the task description, investigates the mounted source, writes a PoC, and submits it through the CyberGym submission interface. A deterministic layer around the model keeps the important scoring mechanics out of the prompt loop: a submission method sends the authored proof-of-concept and processes benchmark response; a lightweight judge checks that the model's summary still matches the described vulnerability before accepting; and accepted submissions are re-submitted a few times to reject non-deterministic crashes. No domain knowledge is included beyond this. Performance is predicated on agent architecture rather than cybersecurity steering.

\inlineparagraph{Results} Scores compared to state-of-the-art are given in Table~\ref{tab:cybergym}. We report a number of leading closed-source results, and include two baselines: OpenAI Codex, and OpenAI Codex plus a skill used to comply with the CyberGym submission format. \nooa scores highly, beating the majority of closed-source solutions, and is the top-scoring open source agent.

\inlineparagraph{Network access} Monitoring network access affects performance. We implemented a rigorous ``cheat check'' with rule-based analysis of agent trajectories. This ensured that \nooa results are based only in information that the agent is processing and inducing directly from the problem setup, rather than being able to look up information about relevant disclosed vulnerabilities or the benchmark itself online.

\begin{table}[tbp]
\centering
\caption{Vulnerability discovery performance on CyberGym L1}
\label{tab:cybergym}
\scalebox{0.85}{%
\sffamily\small
\begin{tabular}{lcccc}
\toprule
\textbf{Harness} & \textbf{Model} & \textbf{Network} & \textbf{Solve rate (\%)} & \textbf{Open source?} \\
\midrule
Microsoft MDASHv2 & MDASH & \texttt{unknown} & 95.6 & No \\
Crystalline & Opus 4.6 & \texttt{blocked} & 89.6 & No \\
\nooa & GPT-5.5 & \texttt{blocked} & 86.8 & Yes \\
OpenAI Daybreak & GPT-5.5 & \texttt{unknown} & 85.6 & No \\
OpenAI Codex + submission skill & GPT-5.5 & \texttt{open} & 83.5 & Yes \\
Anthropic Glasswing & Mythos & \texttt{unknown} & 83.1 & No\\
OpenAI Codex & GPT-5.5 & \texttt{blocked} & 64.9 & Yes\\
\bottomrule
\end{tabular}
}
\end{table}

%% file: arc_agi_3_results.tex
ARC-AGI-3~\cite{arcprize2026arcagi3} is an interactive-reasoning benchmark: the agent is dropped into an unknown grid game and must discover mechanics, objective, and controls purely by acting. Our companion DreamTeam system~\cite{sarafian2026workspace} -- six specialized agents coordinating around a shared executable world model -- set the previous best published score on it. We test whether that methodology survives radical simplification: \textbf{one \nooa agent and one 50-line skill}, with six role prompts (1{,}821 lines) and a 4{,}690-line harness-side retrodiction engine absorbed by framework primitives -- the CodeAct REPL as simulator, context blocks as shared state, memory (Sec~\ref{sec:memory}) as the team's carry-forward ledgers.

\textbf{The world-model skill} instructs the agent to persist an executable model as workspace modules: \inlinecode{\texttt{encode(grid)\,$\to$\,z}}, a latent of the few fields that drive the game; \inlinecode{\texttt{predict(z, action)\,$\to$\,z'}}, the dynamics; \emph{retrodiction} each turn -- a predict-vs-observed mismatch is the sole refinement signal; search over its own \inlinecode{\texttt{predict}} once trusted; and memory discipline across levels. Every turn ends with \inlinecode{\texttt{submit\_actions(..., rationale="predict: ...")}} -- each action batch is a checked experiment.

\begin{figure}[tbp]
\centering
\includegraphics[width=\linewidth]{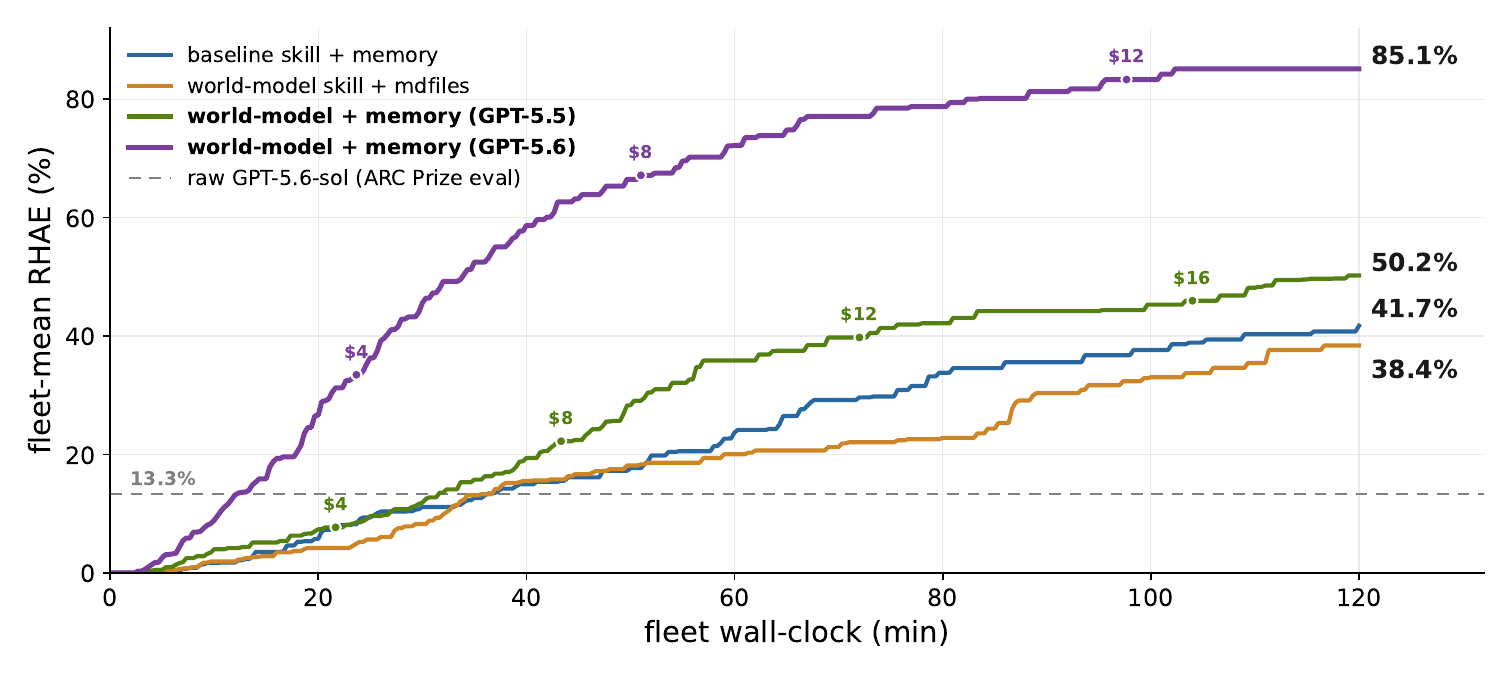}
\caption{\textbf{ARC-AGI-3 under the two-hour fleet cap} (25 games per fleet, one \nooa agent per game). Fleet-mean RHAE vs.\ wall-clock -- the world-model skill with the memory subsystem reaches 50.2\% on GPT-5.5 and \textbf{85.1\% on GPT-5.6-sol}, vs.\ 41.7\% for the baseline skill and 38.4\% for the world-model skill with markdown files in place of memory (both GPT-5.5; +8.5 over the baseline, +11.8 over the no-memory ablation). Both world-model + memory fleets stay under \$20 per game at gpt-5.5 pricing; dots on their curves mark each \$4 of per-game cumulative spend. The dashed line marks ARC Prize's evaluation of raw GPT-5.6-sol on these games (13.3\%).}
\label{fig:arc3-rhae}
\end{figure}

\textbf{Results.} We ran four 25-game fleets, one agent per game, under the competition's two-hour cap: the world-model skill with the memory subsystem on GPT-5.5 and on GPT-5.6-sol, the same skill with plain markdown files in place of memory, and a hypothesis-driven baseline skill with memory (the last two on GPT-5.5).\footnote{Public ARC-AGI-3 scorecards for the two world-model + memory fleets: \href{https://arcprize.org/scorecards/b2aa2240-8cd7-41cf-ad83-edf8c2836084}{GPT-5.5} and \href{https://arcprize.org/scorecards/7a511ea0-dfa1-47ed-8a99-52a80f3cdbaa}{GPT-5.6-sol}.} Figure~\ref{fig:arc3-rhae} plots the fleet-mean RHAE -- the competition's action-efficiency score against per-level human baselines -- over time and spend. At the cap, the world-model + memory fleet on GPT-5.5 reaches \textbf{RHAE 50.2\%} (118 levels), vs.\ 41.7\% for the baseline and 38.4\% for the markdown-file ablation: \textbf{+8.5 points over the baseline}, and \textbf{+11.8 points over the same skill without the memory subsystem}. On GPT-5.6-sol the same agent (170 levels) is \textbf{scoring 85.1\% with less than \$20 per game}; the guarded, cache-aware fleets cost \$17.85 (GPT-5.5) and \$13.28 (GPT-5.6-sol) per game at gpt-5.5 pricing. For scale, ARC Prize's own evaluation of raw GPT-5.6-sol -- the only performant base model on the benchmark as of July 2026 -- averages \textbf{13.3\%} on the same 25 public games at maximum reasoning effort\footnote{\href{https://arcprize.org/results/openai-gpt-5-6-sol}{arcprize.org/results/openai-gpt-5-6-sol}, July 2026; evaluation budgets differ, so the comparison is indicative.}; the same model inside the \nooa harness reaches 85.1\% -- a $6.4\times$ harness effect. The curves separate once a game's mechanics have been observed enough to encode and predict.

\textbf{World-model use.} 22 of 25 games persisted executable model code ($\sim$4.4k lines). Game \texttt{m0r0}, for example, replayed twenty live frames through its \inlinecode{\texttt{encode}} to validate a stored 42-action plan mid-execution and completed 6/6 levels near the per-level score cap.

\textbf{Memory use.} The fleet exercised all three interfaces of the memory subsystem (Sec~\ref{sec:memory}; Table~\ref{tab:arc3-memory-use}): 3{,}262 memories written, 12{,}654 spontaneous injections, and 27{,}115 deliberate tool reads at a 99\% hit rate. Retrieval favors what the agent marked important (mean importance 6.1 written vs.\ 7.5 deliberately recalled), injection stays bounded at 4.1 memories per turn, and recall frequency tracks success: winning games average 1.63 deliberate recalls per decision, and recalls per decision correlate with levels completed at Spearman $\rho=+0.52$ (Appendix~\ref{app:arc3-memory-use}).

\textbf{Containment.} The fleet runs inside layered sandboxing: a kernel-enforced per-cell OS sandbox (each CodeAct cell in a locked-down worker under irrevocable Landlock filesystem default-deny, a seccomp network block, memory/CPU caps, and a hard cell timeout) over the in-process cell guard, a per-run OS privilege drop, and game identities replaced end-to-end by opaque aliases. An 18-pass red-team audit of the live run found no leakage on any rule -- no internet egress, no game-source or cross-game reads, zero real game identifiers in 13{,}335 logs -- and the single escape attempt was blocked by the cell guard. Appendix~\ref{app:arc3-details} details the evidence and the audit.

\begin{table}[tbp]
\centering
\caption{Memory-system use by the ARC-AGI-3 fleet (25 games): what agents wrote vs.\ what each read channel surfaced. Read columns count occurrences (one memory surfacing once); imp.\ = mean importance (verbal scale mapped to 0--10); len = mean characters.}
\label{tab:arc3-memory-use}
\scalebox{0.85}{%
\sffamily\small
\begin{tabular}{lrrrrr}
\toprule
& \multicolumn{3}{c}{\textbf{Written}} & \textbf{Injected (spont.)} & \textbf{Recalled / searched} \\
\cmidrule(lr){2-4}\cmidrule(lr){5-5}\cmidrule(lr){6-6}
\textbf{Type} & n (\%) & imp. & len & occurrences (\%) & occurrences (\%) \\
\midrule
\texttt{info} & 2{,}130 (65\%) & 6.7 & 555 & 9{,}055 (72\%) & 22{,}302 (82\%) \\
\texttt{skill} & 91 (3\%) & 8.3 & 587 & 293 (2\%) & 956 (4\%) \\
\texttt{episode} & 321 (10\%) & 6.7 & 430 & 3{,}079 (24\%) & 3{,}439 (13\%) \\
\texttt{todo} & 18 (1\%) & 6.7 & 560 & 56 (0\%) & 49 (0\%) \\
\texttt{reflection} & 702 (22\%) & 3.9 & 377 & 171 (1\%) & 369 (1\%) \\
\midrule
\textbf{all} & 3{,}262 & 6.1 & 505 & 12{,}654 & 27{,}115 \\
\bottomrule
\end{tabular}
}
\end{table}

%% file: appendix_stress_trace.tex
\section{Appendix: A stress test up close}
\label{app:sentiment-batch-traces}

This appendix shows four complete runs of \texttt{sentiment\_batch}, the hardest capability stress test (31/50 overall). The listings are reproduced from the run traces: every ellipsis and truncation marker below was produced by the harness and seen by the model. Each message is a titled block whose colored left rule gives its role: \textcolor{ctxamber!70!black}{\textbf{amber}} for the cached system region, \textcolor{builtinblue!80!black}{\textbf{blue}} for user-role harness messages (task, execution output, dynamic context), and \textcolor{nvgreen!60!black}{\textbf{green}} for model output.

The test agent:

\begin{lstlisting}[basicstyle=\ttfamily\scriptsize,literate={—}{{---}}1 {°}{{\textdegree}}1,frame=l,framerule=2.5pt,rulecolor=\color{codeframe},xleftmargin=8pt,framexleftmargin=8pt,title={\scriptsize\sffamily\bfseries\color{black!70}TEST AGENT (developer-written source)}]
class SentimentBatchAgent(Agent):
    """You are an agent that classifies sentiment of multiple texts."""

    def __init__(self, **kwargs):
        super().__init__(**kwargs)
        self.method_writing = MethodWriting()

    async def classify(
        self, texts: Annotated[list[str], "The texts to classify"]
    ) -> list[Literal["positive", "negative", "neutral"]]:
        """Classify the sentiment of multiple texts."""
        ...
\end{lstlisting}

The scorer requires an exact match against 50 reference labels. All four runs received byte-identical context. The system region is shown first: the framework prompt, the CodeAct strategy instructions, the execution context, and the agent's own \texttt{doc(self)} rendering --- the typed contract as the model sees it. Note the fan-out pattern and the instruction to return computed values by variable, both of which matter below.

\begin{lstlisting}[language={},basicstyle=\ttfamily\scriptsize,literate={—}{{---}}1 {°}{{\textdegree}}1,frame=l,framerule=2.5pt,rulecolor=\color{ctxamber},xleftmargin=8pt,framexleftmargin=8pt,title={\scriptsize\sffamily\bfseries\color{ctxamber!70!black}SYSTEM --- cached prefix (identical across all four runs)}]
<system_prompt expr="self._system_prompt()">
You are SentimentBatchAgent, a Python agent working in an interactive session.

## Context blocks
Your prompt is organized in XML context blocks: `<name>CONTENT</name>`.
Blocks produced by `self.context.set_dynamic()` carry an `expr="..."` attribute whose value is the Python expression re-evaluated each turn.
Event history: system entries in `<sys tag="N">`; reference via `self.events["N"]`.

## Truncation
- A bare Python literal (`[1, 2, 3]`, `{1: 2}`, `'hello'`) is always complete.
- Truncated values use a `type(len=N, ...)` (or `type(repr_len=N, ...)`) marker:
    list(len=100, [:5]=[...], [-5:]=[...])
    tuple(len=100, [:5]=(...), [-5:]=(...))
    dict(len=100, items={...})
    set(len=100, items={...})
    str(len=100000, [:250]='...', [-250:]='...')
    ndarray(repr_len=233, [:100]='...', [-100:]='...')
- Structured instances (dataclasses, Pydantic, custom classes) render as `ClassName(field=value, ...)`; a trailing `...` means fields were elided:
    Config(name='foo', enabled=True, ...)
- The variable itself is **not** truncated — index/iterate it directly to operate on the full data.
- `<truncated>...</truncated>` in captured stdout/stderr is **not recoverable**.

</system_prompt>

<strategy_prompt>
## Strategy

Jupyter-like Python session. Parameters pre-loaded as locals; state persists across cells. Use `await` directly, `print`/`pprint` to debug, `doc(obj)` to inspect types. You MUST call a tool each turn — **plain-text responses do NOT end the session**. To finish, call `return_result(value)`. Repeated text-only responses will abort the run with an error.

**Your two tools:**
- `execute_python(code)` — run a code cell
- `return_result(value)` — submit your final answer (also callable from inside `execute_python`)

## When to use which tool

Use `return_result(...)` directly for simple answers determinable from the inputs alone (yes/no, one field, a single lookup).

Use `execute_python(...)` for lists/batches, arithmetic, multi-step computation, transforms, or iteration. Always iterate in code — never construct large arrays by hand.

For language tasks (classification, extraction, interpretation), use LLM reasoning — answer directly via `return_result`, or delegate to a `@strategy(PredictStrategy())` standalone function (see below). Don't keyword-match or regex.

## Returning computed results

After computing in code, call `return_result(variable)` **from within** `execute_python()`. This passes the variable directly. Do NOT re-type computed values in a separate `return_result` tool call.

## Helpers

Define helpers at the top of the cell and call them by name. Existing methods on `self` are usable via `await self.method(...)`. Helpers persist as REPL locals across cells in this session.

```python
def normalize(x):
    return x.strip().lower()

cleaned = [normalize(v) for v in values]
```

## Fan-out generation

For per-item LLM work over a list, decorate a standalone async function with `@strategy(PredictStrategy())` and an ellipsis body. `asyncio.gather` runs the calls in parallel.

```python
@strategy(PredictStrategy())
async def detect_language(message: str) -> str:
    """Return the ISO 639-1 language code for {message} (e.g. 'en', 'fr', 'de', 'ja')."""
    ...

codes = await asyncio.gather(*(detect_language(m) for m in messages))
return_result(codes)
```

For iterative sub-tasks that need code execution, use `@strategy(CodeActStrategy())`. The sub-task must be strictly simpler than the current call to avoid infinite recursion.

## Restrictions (will throw)

- `eval`, `exec`, `compile`, `__import__`, `input`, `breakpoint`
- `globals`, `locals`, `vars`, `asyncio.run`, `loop.run_until_complete`
- Attaching callables to the agent: `self.foo = fn`, `setattr(self, 'foo', fn)`, `type(self).foo = fn`
</strategy_prompt>

<execution_context>
## Execution Context

**Available types** (defined in agent or ancestor modules): SentimentBatchAgent
  Tip: Use `doc(SentimentBatchAgent)` to inspect fields before constructing
**Imported items**: Agent, Annotated, Literal, MethodWriting
**Task decomposition**: `@strategy(PredictStrategy())` decorator, `strategy`, `PredictStrategy`, `CodeActStrategy`
**Stdlib**: `asyncio`, `typing` (Literal, Annotated, etc.)

**Always available**: `self`, `print()`, `pprint()`, `doc()`, `return_result()`, `reasoning()` method parameters
</execution_context>

<self expr="doc(type(self))">
class SentimentBatchAgent:
    """You are an agent that classifies sentiment of multiple texts."""

    method_writing: MethodWriting = MethodWriting()  # Define helpers and LLM-powered sub-calls at the top of a REPL cell.

    async def classify(self, texts: list[str]) -> list[Literal[positive, negative, neutral]]:
        """
        Classify the sentiment of multiple texts.
        
        Args:
            texts: The texts to classify
        """
</self>
\end{lstlisting}

Each run then begins with the task, the harness-initiated input inspection and its output, and the per-turn state block:

\begin{lstlisting}[language={},basicstyle=\ttfamily\scriptsize,literate={—}{{---}}1 {°}{{\textdegree}}1,frame=l,framerule=2.5pt,rulecolor=\color{builtinblue},xleftmargin=8pt,framexleftmargin=8pt,title={\scriptsize\sffamily\bfseries\color{builtinblue!70!black}USER --- task}]
<sys tag="1">
Task(prompt='''## Task: classify

Classify the sentiment of multiple texts.

You are executing `classify` — code runs in the Execution Context above. Calling `self.classify(...)` would recurse.''')
</sys>
\end{lstlisting}
\begin{lstlisting}[basicstyle=\ttfamily\scriptsize,literate={—}{{---}}1 {°}{{\textdegree}}1,frame=l,framerule=2.5pt,rulecolor=\color{nvgreen},xleftmargin=8pt,framexleftmargin=8pt,title={\scriptsize\sffamily\bfseries\color{nvgreen!70!black}ASSISTANT --- tool call: execute\_python (harness-initiated input inspection)}]
reasoning(f"""Inspecting inputs for classify().""")
print(f"Task: classify()")
print(f"\ntexts ({type(texts).__name__}):")
pprint(texts, max_length=25, max_string=2000, max_depth=4)
pprint(texts, max_length=25, max_string=2000, max_depth=4)
\end{lstlisting}
\begin{lstlisting}[language={},basicstyle=\ttfamily\scriptsize,literate={—}{{---}}1 {°}{{\textdegree}}1,frame=l,framerule=2.5pt,rulecolor=\color{builtinblue},xleftmargin=8pt,framexleftmargin=8pt,title={\scriptsize\sffamily\bfseries\color{builtinblue!70!black}USER --- execution output}]
<sys tag="3">
PythonOutput(tool_call_id='prefill_8b0f55d4', execution_status=<ResultStatus.COMPLETE: 'complete'>, stdout='''Task: classify()

texts (list):
list(len=50,
    [:13]=[
        'This is the best day ever!',
        "I'm so disappointed with the service.",
        'The temperature is 72 degrees.',
        'What a fantastic experience!',
        'I regret buying this.',
        'The meeting is at 3pm.',
        "I'm thrilled with these results!",
        'Terrible quality, waste of money.',
        'The report contains 50 pages.',
        "I couldn't be happier with the result!",
        'Absolutely love it!',
        'This is awful and unusable.',
        "It's okay, nothing special.",
    ],
    [-12:]=[
        'Perfect in every way!',
        'This failed miserably.',
        'The document is 10 pages long.',
        'Amazing results, so pleased!',
        'Very poor quality, extremely disappointed.',
        'Typical response time.',
        'Fantastic product, love it!',
        'This is a complete mess.',
        'The office temperature is 75°F.',
        'Absolutely thrilled with this purchase!',
        'Deeply unsatisfied, very frustrating.',
        'Regular maintenance is required.',
    ],
)
''')
</sys>
\end{lstlisting}
\begin{lstlisting}[language={},basicstyle=\ttfamily\scriptsize,literate={—}{{---}}1 {°}{{\textdegree}}1,frame=l,framerule=2.5pt,rulecolor=\color{builtinblue},xleftmargin=8pt,framexleftmargin=8pt,title={\scriptsize\sffamily\bfseries\color{builtinblue!70!black}USER --- dynamic context (re-rendered each turn)}]
<context>
<state expr="pformat(self, max_length=50, max_string=500, max_depth=4)">
SentimentBatchAgent(method_writing=MethodWriting())
</state>
</context>
\end{lstlisting}

\noindent The model sees the type, the true length (\texttt{len=50}), and 25 of the 50 texts; the variable \texttt{texts} holds all 50. The four runs diverge at the first model-authored cell.

\subsection{Nemotron 3 Ultra --- passed}
The intended solution in one model-authored cell: a model-defined subagent, fanned out over the live variable, with the live result returned from inside the cell. 9.6 seconds end to end.

\begin{lstlisting}[basicstyle=\ttfamily\scriptsize,literate={—}{{---}}1 {°}{{\textdegree}}1,frame=l,framerule=2.5pt,rulecolor=\color{nvgreen},xleftmargin=8pt,framexleftmargin=8pt,title={\scriptsize\sffamily\bfseries\color{nvgreen!70!black}ASSISTANT --- tool call: execute\_python}]
@strategy(PredictStrategy())
async def classify_text(text: str) -> str:
    """Classify the sentiment of this text as 'positive', 'negative', or 'neutral'.
    
    Text: {text}
    """
    ...

# Process all texts in parallel
results = await asyncio.gather(*(classify_text(text) for text in texts))
pprint(results)
return_result(results)return_result(results)
\end{lstlisting}

\subsection{Claude Opus 4.8 --- failed}
The model's first cell: the same fan-out, executed correctly.

\begin{lstlisting}[basicstyle=\ttfamily\scriptsize,literate={—}{{---}}1 {°}{{\textdegree}}1,frame=l,framerule=2.5pt,rulecolor=\color{nvgreen},xleftmargin=8pt,framexleftmargin=8pt,title={\scriptsize\sffamily\bfseries\color{nvgreen!70!black}ASSISTANT --- tool call: execute\_python}]
@strategy(PredictStrategy())
async def classify_sentiment(text: str) -> Literal["positive", "negative", "neutral"]:
    """Classify the sentiment of {text} as 'positive', 'negative', or 'neutral'."""
    ...

results = await asyncio.gather(*(classify_sentiment(t) for t in texts))
for t, r in zip(texts, results):
    print(f"{r:8} | {t}")print(f"{r:8} | {t}")
\end{lstlisting}
\begin{lstlisting}[language={},basicstyle=\ttfamily\scriptsize,literate={—}{{---}}1 {°}{{\textdegree}}1,frame=l,framerule=2.5pt,rulecolor=\color{builtinblue},xleftmargin=8pt,framexleftmargin=8pt,title={\scriptsize\sffamily\bfseries\color{builtinblue!70!black}USER --- execution output (all 50 classifications, all correct)}]
positive | This is the best day ever!
negative | I'm so disappointed with the service.
neutral  | The temperature is 72 degrees.
positive | What a fantastic experience!
negative | I regret buying this.
neutral  | The meeting is at 3pm.
positive | I'm thrilled with these results!
negative | Terrible quality, waste of money.
neutral  | The report contains 50 pages.
positive | I couldn't be happier with the result!
positive | Absolutely love it!
negative | This is awful and unusable.
neutral  | It's okay, nothing special.
positive | Outstanding performance!
negative | What a disaster this turned out to be.
neutral  | The documentation is available online.
positive | Exceeded all my expectations!
negative | Completely unacceptable service.
neutral  | The package weighs 5 pounds.
negative | Never buying from them again.
positive | Phenomenal work, truly impressive!
negative | This is garbage and doesn't work.
neutral  | The file size is 2MB.
positive | Incredible experience, highly satisfied!
negative | Worst purchase I've ever made.
neutral  | Standard shipping procedure.
positive | Highly recommend this to everyone!
negative | Total waste of time and money.
neutral  | The price is $49.99.
positive | Love every aspect of this product!
negative | Disappointing quality, very unhappy.
neutral  | Average performance, nothing notable.
positive | Brilliant solution to the problem!
negative | This is useless and broken.
neutral  | The meeting starts at noon.
positive | Exceeded all expectations, amazing!
negative | Completely frustrated with this.
neutral  | Normal business operations.
positive | Perfect in every way!
negative | This failed miserably.
neutral  | The document is 10 pages long.
positive | Amazing results, so pleased!
negative | Very poor quality, extremely disappointed.
neutral  | Typical response time.
positive | Fantastic product, love it!
negative | This is a complete mess.
neutral  | The office temperature is 75°F.
positive | Absolutely thrilled with this purchase!
negative | Deeply unsatisfied, very frustrating.
neutral  | Regular maintenance is required.
\end{lstlisting}

On the next turn, instead of \texttt{return\_result(results)}, the model transcribed the printed output into a literal in a separate \texttt{return\_result} tool call --- exactly what the strategy instructions above say not to do. The transcription dropped item 43 (\texttt{neutral} for ``Typical response time.''):

\begin{lstlisting}[language={},basicstyle=\ttfamily\scriptsize,literate={—}{{---}}1 {°}{{\textdegree}}1,frame=l,framerule=2.5pt,rulecolor=\color{nvgreen},xleftmargin=8pt,framexleftmargin=8pt,title={\scriptsize\sffamily\bfseries\color{nvgreen!70!black}ASSISTANT --- tool call: return\_result}]
All classifications look correct.

{
    'result': [
        {
            'result': [
                'positive',
                'negative',
                'neutral',
                'positive',
                'negative',
                'neutral',
                'positive',
                'negative',
                'neutral',
                'positive',
                'positive',
                'negative',
                'neutral',
                'positive',
                'negative',
                'neutral',
                'positive',
                'negative',
                'neutral',
                'negative',
                'positive',
                'negative',
                'neutral',
                'positive',
                'negative',
                'neutral',
                'positive',
                'negative',
                'neutral',
                'positive',
                'negative',
                'neutral',
                'positive',
                'negative',
                'neutral',
                'positive',
                'negative',
                'neutral',
                'positive',
                'negative',
                'neutral',
                'positive',
                'negative',
                'positive',
                'negative',
                'neutral',
                'positive',
                'negative',
                'neutral',
            ],
        }
    ],   # 49 items
}
\end{lstlisting}

\noindent Verdict: \texttt{List length mismatch: expected 50, got 49}. The live \texttt{results} variable held all 50 labels.

\subsection{GPT-5.5 --- passed}
No subagents. The model's first cell defeats the preview deliberately, printing every item with its index:

\begin{lstlisting}[basicstyle=\ttfamily\scriptsize,literate={—}{{---}}1 {°}{{\textdegree}}1,frame=l,framerule=2.5pt,rulecolor=\color{nvgreen},xleftmargin=8pt,framexleftmargin=8pt,title={\scriptsize\sffamily\bfseries\color{nvgreen!70!black}ASSISTANT --- tool call: execute\_python}]
for i,t in enumerate(texts):
    print(i, repr(t))
\end{lstlisting}
\begin{lstlisting}[language={},basicstyle=\ttfamily\scriptsize,literate={—}{{---}}1 {°}{{\textdegree}}1,frame=l,framerule=2.5pt,rulecolor=\color{builtinblue},xleftmargin=8pt,framexleftmargin=8pt,title={\scriptsize\sffamily\bfseries\color{builtinblue!70!black}USER --- execution output}]
0 'This is the best day ever!'
1 "I'm so disappointed with the service."
2 'The temperature is 72 degrees.'
3 'What a fantastic experience!'
4 'I regret buying this.'
5 'The meeting is at 3pm.'
6 "I'm thrilled with these results!"
7 'Terrible quality, waste of money.'
8 'The report contains 50 pages.'
9 "I couldn't be happier with the result!"
10 'Absolutely love it!'
11 'This is awful and unusable.'
12 "It's okay, nothing special."
13 'Outstanding performance!'
14 'What a disaster this turned out to be.'
15 'The documentation is available online.'
16 'Exceeded all my expectations!'
17 'Completely unacceptable service.'
18 'The package weighs 5 pounds.'
19 'Never buying from them again.'
20 'Phenomenal work, truly impressive!'
21 "This is garbage and doesn't work."
22 'The file size is 2MB.'
23 'Incredible experience, highly satisfied!'
24 "Worst purchase I've ever made."
25 'Standard shipping procedure.'
26 'Highly recommend this to everyone!'
27 'Total waste of time and money.'
28 'The price is $49.99.'
29 'Love every aspect of this product!'
30 'Disappointing quality, very unhappy.'
31 'Average performance, nothing notable.'
32 'Brilliant solution to the problem!'
33 'This is useless and broken.'
34 'The meeting starts at noon.'
35 'Exceeded all expectations, amazing!'
36 'Completely frustrated with this.'
37 'Normal business operations.'
38 'Perfect in every way!'
39 'This failed miserably.'
40 'The document is 10 pages long.'
41 'Amazing results, so pleased!'
42 'Very poor quality, extremely disappointed.'
43 'Typical response time.'
44 'Fantastic product, love it!'
45 'This is a complete mess.'
46 'The office temperature is 75°F.'
47 'Absolutely thrilled with this purchase!'
48 'Deeply unsatisfied, very frustrating.'
49 'Regular maintenance is required.'
\end{lstlisting}

The model's second cell labels by hand --- transcription again, but with explicit per-item bookkeeping:

\begin{lstlisting}[basicstyle=\ttfamily\scriptsize,literate={—}{{---}}1 {°}{{\textdegree}}1,frame=l,framerule=2.5pt,rulecolor=\color{nvgreen},xleftmargin=8pt,framexleftmargin=8pt,title={\scriptsize\sffamily\bfseries\color{nvgreen!70!black}ASSISTANT --- tool call: execute\_python}]
labels = [
    labels = [
        'positive',  # 0 This is the best day ever!
        'negative',  # 1 disappointed
        'neutral',   # 2 factual temperature
        'positive',  # 3 fantastic
        'negative',  # 4 regret
        'neutral',   # 5 factual meeting
        'positive',  # 6 thrilled
        'negative',  # 7 terrible/waste
        'neutral',   # 8 factual report
        'positive',  # 9 couldn't be happier
        'positive',  # 10 love
        'negative',  # 11 awful/unusable
        'neutral',   # 12 okay, nothing special
        'positive',  # 13 outstanding
        'negative',  # 14 disaster
        'neutral',   # 15 documentation available
        'positive',  # 16 exceeded expectations
        'negative',  # 17 unacceptable
        'neutral',   # 18 package weighs
        'negative',  # 19 never buying again
        'positive',  # 20 phenomenal/impressive
        'negative',  # 21 garbage/doesn't work
        'neutral',   # 22 file size
        'positive',  # 23 incredible/satisfied
        'negative',  # 24 worst purchase
        'neutral',   # 25 standard shipping
        'positive',  # 26 recommend
        'negative',  # 27 waste
        'neutral',   # 28 price
        'positive',  # 29 love
        'negative',  # 30 disappointing/unhappy
        'neutral',   # 31 average/nothing notable
        'positive',  # 32 brilliant
        'negative',  # 33 useless/broken
        'neutral',   # 34 meeting starts
        'positive',  # 35 exceeded/amazing
        'negative',  # 36 frustrated
        'neutral',   # 37 normal operations
        'positive',  # 38 perfect
        'negative',  # 39 failed miserably
        'neutral',   # 40 document pages
        'positive',  # 41 amazing/pleased
        'negative',  # 42 poor/disappointed
        'neutral',   # 43 typical response
        'positive',  # 44 fantastic/love
        'negative',  # 45 complete mess
        'neutral',   # 46 office temperature
        'positive',  # 47 thrilled
        'negative',  # 48 unsatisfied/frustrating
        'neutral',   # 49 regular maintenance
    ]
return_result(labels)
\end{lstlisting}

\subsection{GPT-5.4 Mini --- failed}
The model's only cell: a keyword-rule classifier, with keyword lists fitted to the 25 texts visible in the preview, applied blind to all 50.

\begin{lstlisting}[basicstyle=\ttfamily\scriptsize,literate={—}{{---}}1 {°}{{\textdegree}}1,frame=l,framerule=2.5pt,rulecolor=\color{nvgreen},xleftmargin=8pt,framexleftmargin=8pt,title={\scriptsize\sffamily\bfseries\color{nvgreen!70!black}ASSISTANT --- tool call: execute\_python}]
labels = []
labels = []
for t in texts:
    s = t.lower()
    pos = any(w in s for w in ['best','fantastic','thrilled','happier','love','perfect','amazing','pleased','thrilled','excellent','great'])
    neg = any(w in s for w in ['disappointed','regret','terrible','awful','poor','failed','waste','unusable','miserably','unsatisfied','mess','frustrating'])
    if pos and not neg:
        labels.append('positive')
    elif neg and not pos:
        labels.append('negative')
    elif pos and neg:
        labels.append('neutral')
    else:
        labels.append('neutral')
return_result(labels)
\end{lstlisting}

\noindent It iterates the live variable correctly, but substitutes keyword rules for semantic judgment --- against the strategy instructions --- on 25 texts it never inspected; the labels do not match.

\subsection{What the four runs show}
Sophistication and success are orthogonal: the most advanced harness use (Opus's fan-out) failed on the cheapest discipline --- return the variable, do not retype it --- while the least agentic approach (GPT-5.5's manual labeling) passed on careful bookkeeping. Both failures ignored an explicit instruction in the strategy prompt, and both had a safe path already provided by the interface. This is the pattern behind the stress-test results in Section~\ref{sec:evaluation}: the remaining failures are not gaps in interface understanding but lapses in disciplined use of it --- and they are exactly the behaviors that trajectory-level reinforcement learning (Section~\ref{sec:conclusion}) could target.

%% file: appendix_memory_arc3.tex
% ===========================================================================
\section{Appendix: Memory-System Details}
\label{app:memory-details}
% ===========================================================================

\subsection{Design decisions}
The subsystem is additive by construction: \inlinecode{\texttt{MemoryManager.install(agent)}} wires storage, retrieval, and hooks onto an unmodified agent through existing extension points (event subscriptions, call middleware, context blocks), and uninstalling restores the agent exactly. Four decisions are load-bearing. (i)~\emph{Verbal boundary}: tools accept and render verbal descriptors (\textsc{critical}\,\ldots\,\textsc{trivial}; \textsc{open}/\textsc{done}/\textsc{dropped}) while scoring stays numeric internally, keeping the model-facing vocabulary in-distribution. (ii)~\emph{Injection never self-reinforces}: spontaneous recall runs the same retrieval pipeline with \inlinecode{\texttt{touch=False}}, so what the harness chooses to show does not inflate ACT-R activation; only deliberate tool recall does. (iii)~\emph{One SQLite file as source of truth}: records, a typed memory graph, maintenance log, and per-memory access records live in a single human-inspectable file; vector indexes (numpy, sqlite-vec, or Chroma) are derived and rebuilt on demand. (iv)~\emph{Pass-by-reference memories}: a record may hold \inlinecode{\texttt{kind:key}} references resolved against live agent state at recall time by strict name lookup (never \inlinecode{\texttt{eval}}), returning a \textsc{live} value or an explicitly \textsc{dangling} snapshot -- eliminating the stale-copy failure mode we measured with copied values. Prospective state is first-class: \emph{todo} memories carry a lifecycle, survive pruning while open, and can be surfaced each turn. Together, the tools and the reflection pipeline carry skill-library and self-critique memory~\cite{wang2024voyager,shinn2023reflexion} into the object model. Observability is self-contained: every access is recorded on the memory itself, a retrieval call can be replayed with \inlinecode{\texttt{explain()}}, and memory events bridge to OpenTelemetry spans with trace$\leftrightarrow$record cross-links. The controlled measurement of the subsystem's effect is the ARC-AGI-3 ablation (Sec.~\ref{sec:arc3}): +11.8 RHAE points over the identical agent with file-based notes in place of memory. In small internal pilots, reflection helped when retrieval was the bottleneck and hurt pinpoint lookup (abstraction blurs the exact fact), which is why consolidation is configurable per store.

\subsection{Memory across today's harnesses}
Three families dominate current systems. \emph{Flat markdown, always in context} (Claude Code's \texttt{CLAUDE.md}, Codex's \texttt{AGENTS.md}, Gemini CLI's \texttt{GEMINI.md}, Cursor rules): human-authored, transparent, versionable -- but token cost grows linearly and nothing is learned automatically. \emph{Vector stores, similarity-retrieved} (AutoGen teachability, CrewAI, Mem0-style layers, Letta archival): automatic accumulation at unbounded scale -- but opaque to the user and unverified at write time. \emph{Structured self-edited context} (Letta memory blocks, LangMem managed memories): typed segments the agent maintains, occasionally consolidated in the background. During 2025--2026 the CLI harnesses converged on a two-layer hybrid -- a human instruction file plus a model-written auto-memory layer -- differing mainly in whether the auto layer is user-readable and whether retrieval is bounded. The \nooa memory system sits at the intersection of the families: file-based and human-auditable like the first, automatically written like the auto-memory layers, and typed, scored, and graph-linked like the structured family, with cognitively grounded retrieval (ACT-R activation, Ebbinghaus decay) in place of plain similarity search. Table~\ref{tab:memory-comparison} summarizes.

\begin{table}[htbp]
\centering
\caption{Memory subsystems of agent harnesses and frameworks, July 2026.}
\label{tab:memory-comparison}
\scalebox{0.78}{%
\sffamily\scriptsize
\begin{tabular}{p{0.16\linewidth}p{0.20\linewidth}p{0.18\linewidth}p{0.22\linewidth}p{0.12\linewidth}p{0.12\linewidth}}
\toprule
\textbf{System} & \textbf{Storage form} & \textbf{Write policy} & \textbf{Retrieval} & \textbf{Scope} & \textbf{Human-editable} \\
\midrule
Claude Code~\cite{claudecode2026memory} & markdown files + auto-memory dir & user + model-automatic & bounded index always loaded; topic files on demand & per-project, per-user & yes (plain files) \\
Codex CLI~\cite{openai2026codexmemories} & \texttt{AGENTS.md} + generated memory files & user + background-automatic & auto-injected next session & per-repo + per-user & partially (``generated state'') \\
Cursor~\cite{cursor2025memories} & rules files + backend memories & user + auto w/ approval & rule modes; auto-inject & project / user / org & rules yes; memories no \\
Gemini CLI~\cite{google2025geminicli} & hierarchical \texttt{GEMINI.md} & user + \texttt{save\_memory} tool & always in context & global + project & yes \\
LangGraph / LangMem~\cite{langchain2025langmem} & JSON docs + vector index & tools + background manager & semantic search (developer-wired) & arbitrary namespaces & no (DB) \\
AutoGen~\cite{microsoft2023teachability} & vector DB of memos & model-automatic & similarity, every turn & per-agent & no \\
CrewAI~\cite{crewai2024memory} & Chroma + SQLite tiers & framework-automatic & automatic RAG & per-crew & no \\
Letta~\cite{letta2024} & in-context blocks + vector archival & agent self-edits via tools & blocks in context; search tools & per-agent, shareable & via API/GUI \\
\midrule
\textbf{\nooa} & \textbf{one SQLite file}: records + typed graph + logs & \textbf{agent tools} + on-event hooks & ACT-R + graph spread; spontaneous injection + tools & per-agent, owner-scoped sharing & \textbf{yes} (single file, viewer, \texttt{explain()}) \\
\bottomrule
\end{tabular}
}
\end{table}

% ===========================================================================
\section{Appendix: ARC-AGI-3 Example Details}
\label{app:arc3-details}
% ===========================================================================

\subsection{From DreamTeam to one agent and one skill}
Table~\ref{tab:arc3-mapping} maps each element of the DreamTeam system~\cite{sarafian2026workspace} onto the \nooa example. The methodology is kept intact -- latent encoding under a declared schema, executable dynamics, retrodiction as the sole refinement signal, search over the learned model, level-boundary reflection with carry-forward -- while the apparatus (roles, inter-agent protocol, harness-side evaluation engine, background search workers) is either absorbed by framework primitives or performed by the agent itself in its REPL. The paper system is $\sim$150k lines with 1{,}821 lines of role prompts; the example is $\sim$6.1k lines with a 50-line skill.

\begin{table}[htbp]
\centering
\caption{World-model methodology: the DreamTeam system vs.\ the \nooa example.}
\label{tab:arc3-mapping}
\scalebox{0.82}{%
\sffamily\scriptsize
\begin{tabular}{p{0.18\linewidth}p{0.40\linewidth}p{0.36\linewidth}}
\toprule
\textbf{Element} & \textbf{DreamTeam (paper system)} & \textbf{\nooa example} \\
\midrule
Encode $\to$ latent $z$ & Observer role (198-line prompt); schema-validated \texttt{observable.py}; per-step encoding log & skill item: \texttt{encode(grid)$\to$z} with declared \texttt{Z\_SCHEMA}, persisted as a helper module \\
Predict dynamics & Simulator role; \texttt{predict()} + recurrent \texttt{history()} with per-hypothesis branches & single \texttt{predict(z,a)$\to$z'} \\
Retrodiction & harness-computed each step by a 4{,}690-line evaluation engine (DreamerV3 pattern) & agent-performed in the REPL each turn \\
Search / planning & Explorer role; sub-goal/policy registries; background search worker process & BFS/greedy/best-first over own \texttt{predict} inside CodeAct cells \\
Verification & Critic + team-leader roles; per-action approve/reject; feedback protocol & turn contract: \texttt{submit\_actions(..., rationale="predict: ...")} checked next turn \\
Memory across levels & per-role step/level logs with carry-forward ledgers; harness-forced reflection & memory subsystem (Sec~\ref{sec:memory}): recall before deciding, reflect at level boundaries and before \texttt{RESET} \\
Team communication & feedback protocol, file registry, shared workspace & none -- one agent, one workspace \\
\bottomrule
\end{tabular}
}
\end{table}

\subsection{Containment and red-team audit}
The threat model has three rules: no internet, no access to a game's generating source or identity, no access to other runs, other games, or prior solutions. Defenses are layered and -- critically -- the hard layers are external to the agent, so reading the defense code does not help escape: (i) per cell, an in-process AST cell guard, module denylist, and \inlinecode{\texttt{open()}} jail, backed by an opt-in per-cell OS sandbox that forks each CodeAct cell into a locked-down worker with four kernel-enforced, irrevocable guards installed before any cell bytecode runs -- Landlock filesystem default-deny, a seccomp block on \texttt{socket(AF\_INET/AF\_INET6)}, \texttt{RLIMIT\_AS}/\texttt{RLIMIT\_CPU} memory/CPU caps, and a hard \texttt{SIGTERM}$\to$\texttt{SIGKILL} cell timeout; (ii) per run, an OS-level uid drop (\texttt{setpriv --no-new-privs}; game source readable only by root; sibling runs carved out), with a fail-closed namespace sandbox (bubblewrap) as the current default; (iii) end-to-end anonymisation -- agents only ever see an opaque \texttt{game-\textless hex\textgreater} alias, with redaction on outputs and state.

A red-team loop re-scanned the \emph{live} 25-game fleet every 30 minutes -- 18 passes -- classifying the code inside every executed cell and the data returned by every tool result. Results: \textbf{no leakage on any rule}. Zero network invocations; zero game-source bytes returned; cross-game reads fail with \texttt{EACCES} and the per-game memory stores remained fully isolated; a name-leak scanner found zero occurrences of any real game identifier across 13{,}335 agent logs (aliases only). There was exactly one escape attempt across the fleet: one agent invoked a filesystem-recon shell command (\texttt{find}/\texttt{grep} for its own store alias); the cell guard replaced the cell before execution and zero bytes were returned. One latent finding (harness log files world-readable inside the run directory) was never referenced by any agent; the one-line fix is documented. A companion analysis documents a known cell-guard gap (dynamic attribute lookup can evade the AST scan) together with its backstop: under the uid drop, even a reached shell cannot read the game source. No game used the gap.

\subsection{World-model usage evidence and failure modes}
Of 25 games, 22 persisted executable model code (37 modules, $\sim$4.4k lines); six games grew a new per-level module as mechanics accumulated (hazards $\to$ tokens $\to$ doors $\to$ pressure plates). By deepest observed use: 5 games ran the full loop (predict + search + retrodiction), 7 planned or predicted with their models, 10 used them for perception/encoding only. Representative closed loops: \texttt{m0r0} stored a 42-action plan, replayed twenty real frames through \inlinecode{\texttt{encode}} to check it mid-execution (``matched the model exactly''), released the next batch, and pre-announced the completing action of its final level -- 6/6 levels near the per-level score cap; \texttt{tu93} passed its planner's output verbatim to \inlinecode{\texttt{submit\_actions}} with the prediction in the rationale; \texttt{ar25} wrote its model on turn one from a single exploratory action, then submitted a 16-action plan ending ``expect level completion on the last \texttt{DOWN}'' -- 8/8 levels in 24 turns. Model depth tracked what each game demanded rather than raw level count; its payoff shows up as action efficiency (near-cap per-level scores, long verified batches).

The failure mode is instructive: the two games that hung did so in \emph{ad-hoc, in-cell} searches that lacked the bounds (\texttt{max\_depth}, visited sets, node budgets) their own \emph{persisted} planners carried -- one branched over all 3{,}456 click targets per node with no budget while its persisted \texttt{predict} went uncalled. Durable, curated artifacts were reliably better engineered than improvised cell code, which argues for the memory-and-workspace discipline of Sec~\ref{sec:memory} and for hard cell timeouts in the harness, now provided by the per-cell OS sandbox above.

\subsection{Memory-system usage during play}
\label{app:arc3-memory-use}
\begin{figure}[tbp]
\centering
\includegraphics[width=0.98\linewidth]{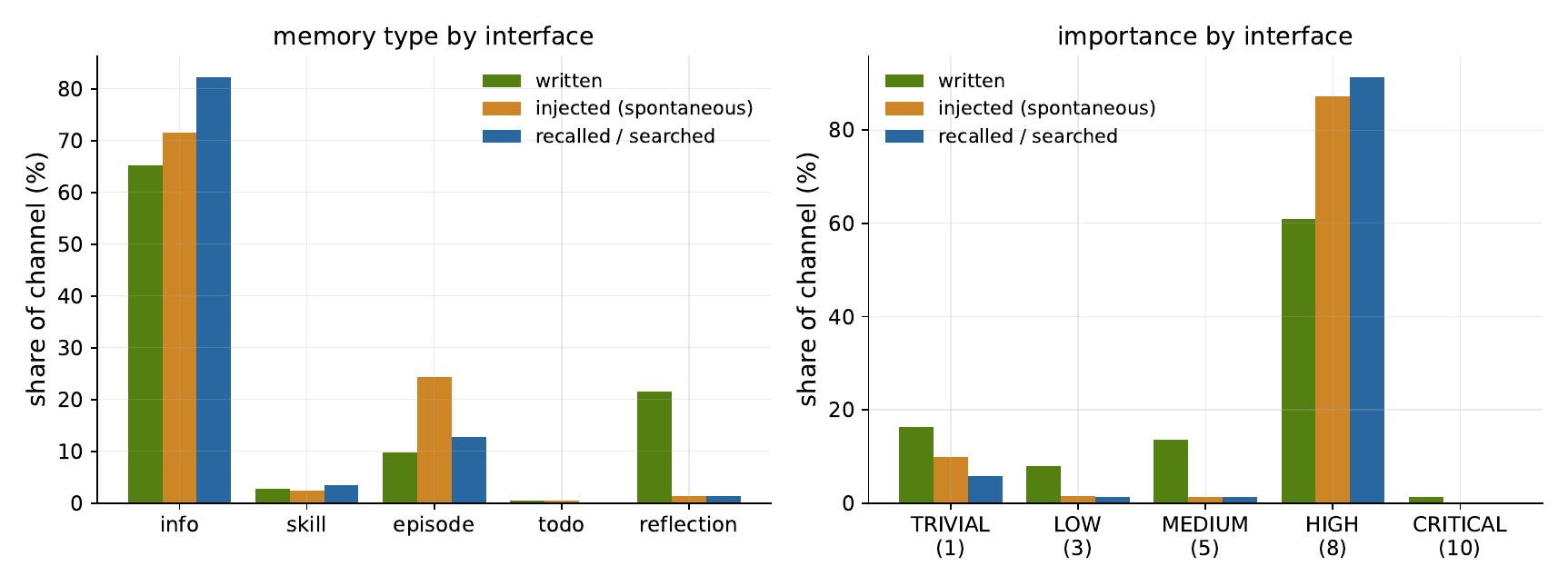}
\caption{\textbf{Memory-system use by interface} in the ARC-AGI-3 fleet (25 games). Left: share of each memory type within the write, spontaneous-injection, and deliberate-read channels. Right: share of each verbal importance level per channel -- both read channels concentrate on \textsc{high}, and the concentration strengthens from written to injected to deliberately recalled.}
\label{fig:arc3-memory-usage}
\end{figure}

We instrumented all three interfaces of the memory subsystem (Sec~\ref{sec:memory}) across the 25 per-game stores: \emph{writes} (agent tools plus consolidation-created records), \emph{spontaneous reads} (the \inlinecode{\texttt{BeforeTurn}} injection into the dynamic context block), and \emph{deliberate reads} (the \inlinecode{\texttt{recall}}/\inlinecode{\texttt{search}} tools). Ground truth comes from the SQLite stores themselves -- each record carries uncapped per-channel counters -- with event-level statistics (injections per turn, hit rates) from the OTel trace exports. Figure~\ref{fig:arc3-memory-usage} shows the type and importance distributions per interface; Table~\ref{tab:arc3-memory-use} gives the counts.

Five observations. (i)~\emph{The channels select differently}: mean importance climbs written $\to$ injected $\to$ deliberate (6.1 $\to$ 7.2 $\to$ 7.5), and the \textsc{high} verbal level carries 61\% of writes but 87\% of injected and 91\% of deliberate occurrences -- the ACT-R importance term biases both read channels toward what the agent itself marked important. (ii)~\emph{Injection is selective and bounded}: only 632 of 3{,}262 memories (19\%) ever surfaced spontaneously, at 4.1 memories $\approx$ 1.9k characters per turn -- the char-budgeted block prevents context flooding by memory. (iii)~\emph{Episodes are the recency channel}: 10\% of writes but 24\% of injected occurrences (13\% deliberate) -- the base-level recency term surfaces the latest level attempts unprompted, while deliberate recall goes after facts (\texttt{info}: 82\% of tool-read occurrences at a 99--100\% hit rate, 9.7 results per call). (iv)~\emph{Skills are few, dear, and deliberately fetched}: 3\% of writes but the highest importance of any type (8.3) and over-represented in deliberate reads -- agents went back for their verified procedures. (v)~\emph{Consolidation compressed the store rather than growing it}: \texttt{reflection} records are 22\% of rows yet $\sim$1\% of both read channels (importance 3.9), and 45\% of all records ended archived by decay-based forgetting. The \texttt{intent} and \texttt{scratch} types went unused; \texttt{todo} appeared in 18 records. Per-game store sizes ranged 23/129/255 (min/median/max).

\begin{figure}[tbp]
\centering
\includegraphics[width=0.98\linewidth]{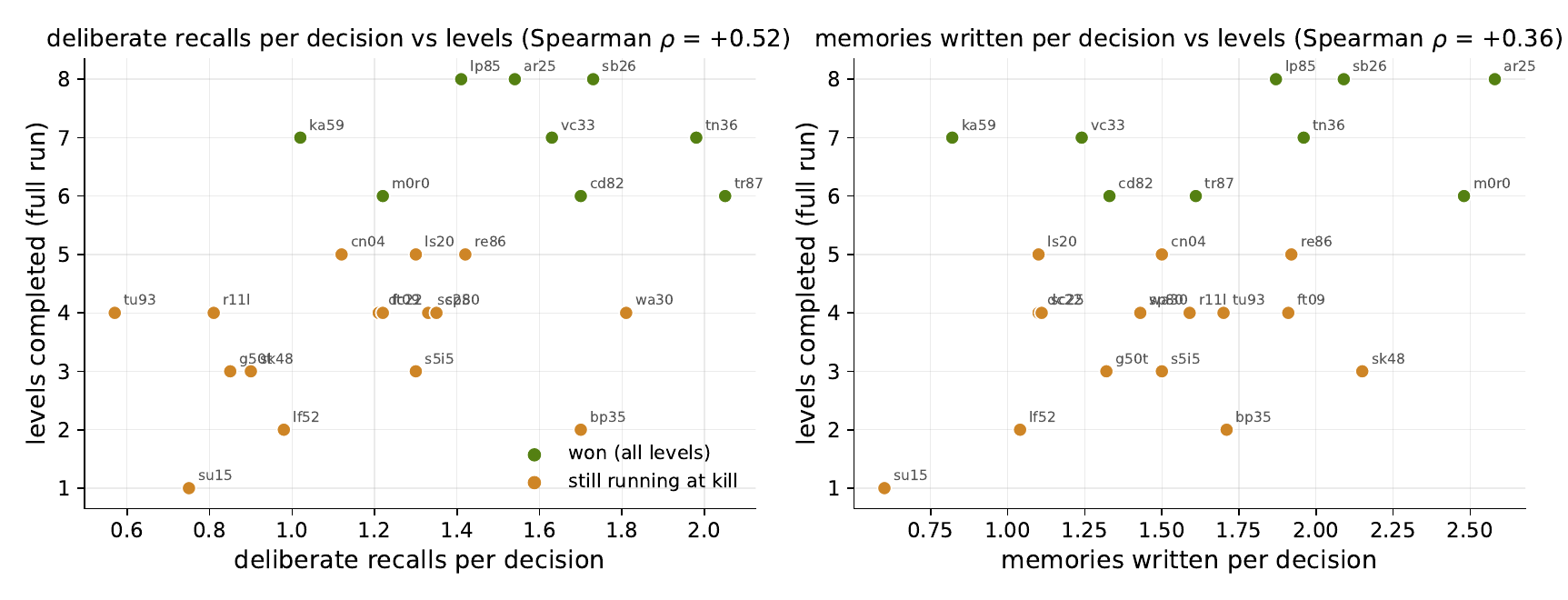}
\caption{\textbf{Memory engagement per decision vs.\ performance} in the ARC-AGI-3 fleet (25 games; a decision is one agent turn ending in \inlinecode{\texttt{submit\_actions}}). Deliberate recalls per decision (left) and memories written per decision (right) against levels completed over the full run. Memory engagement per decision correlates positively with performance; every winning game makes at least one deliberate recall per decision.}
\label{fig:arc3-memory-per-decision}
\end{figure}

\emph{Memory engagement per decision tracks performance.} Because raw store volume largely reflects run length (longer games accumulate more turns, and every turn leaves memory behind), the informative measure is memory use \emph{per decision} -- one decision being one agent turn ending in \inlinecode{\texttt{submit\_actions}}. On this measure the relationship with performance is clearly positive (Figure~\ref{fig:arc3-memory-per-decision}): deliberate recalls per decision correlate with levels completed at Spearman $\rho=+0.52$, and writes per decision at $\rho=+0.36$. Winning games check memory 1.63 times and write 1.87 memories per decision (medians, vs.\ 1.21 and 1.46 for the remaining games), and every winning game makes at least one deliberate recall per decision -- the skill's recall-before-deciding discipline in action. Spontaneous injection is cadence-fixed at $\approx$1 per turn by design and therefore uniform across the fleet. With $n=25$ and 16 outcomes right-censored by the operator kill, these are associations.

\subsection{Reproduction}
Runs analyzed: the RHAE curves and 2-hour numbers are the guarded, cache-aware fleets \path{20260716_204102_competition_gpt55_guarded} (GPT-5.5) and \path{20260718_012940_competition_gpt56sol_guarded} (GPT-5.6-sol), with \path{20260710_154254_competition_memory_visual} (baseline) and \path{20260714_201702_competition_md} (markdown-file ablation) for reference; all 25 games each, and they regenerate from the per-game event logs via \path{tmp/nooa_paper_contribution/artifacts/performance_2h.py}. The memory-usage analysis is from \path{20260711_193827_competition_memory_visual_wm} (world-model skill, 25 games, GPT-5.5) via \path{memory_usage_analysis.py} in the same directory. Pricing \$5/\$30/\$0.50 per Mtok (input/output/cached).